%% file: main.tex
\documentclass[10pt, conference, compsocconf]{IEEEtran}
\input{preamble}

\begin{document}
	
	\title{Detecting Adversarial Examples in Learning-Enabled Cyber-Physical Systems \\ using Variational Autoencoder for Regression
	\thanks{The material presented in this paper is based upon work supported by the National Science Foundation (NSF) under grant numbers CNS 1739328,  the Defense Advanced Research Projects Agency (DARPA) through contract number FA8750-18-C-0089, and the Air Force Office of Scientific Research (AFOSR) DDDAS through contract number FA9550-18-1-0126. The views and conclusions contained herein are those of the authors and should not be interpreted as necessarily representing the official policies or endorsements, either expressed or implied, of AFOSR, DARPA, or NSF.}
	}
	
	\author{\IEEEauthorblockN{Feiyang Cai}
		\IEEEauthorblockA{\textit{Vanderbilt University} \\
			Nashville, TN \\
			feiyang.cai@vanderbilt.edu}
		\and
		\IEEEauthorblockN{Jiani Li}
		\IEEEauthorblockA{\textit{Vanderbilt University} \\
			Nashville, TN \\
			jiani.li@vanderbilt.edu}
		\and
		\IEEEauthorblockN{Xenofon Koutsoukos}
		\IEEEauthorblockA{\textit{Vanderbilt University} \\
			Nashville, TN \\
			xenofon.koutsoukos@vanderbilt.edu}
	}
	
	\maketitle
	\thispagestyle{plain} \pagestyle{plain}
	
	\input{abstract}
	\input{introduction}
	\input{problem}

	\input{method}
	\input{evaluation}

	\input{conclusion}

	\nocite{*}
	\bibliographystyle{IEEEtran}
	\bibliography{references}
	%\begin{thebibliography}  
	%\end{thebibliography}
	
\end{document}

%% file: preamble.tex
\IEEEoverridecommandlockouts
\usepackage{cite}
\usepackage{amsmath,amssymb,amsfonts}
\usepackage{algorithmic}
\usepackage{graphicx}
\usepackage{textcomp}
\usepackage{xcolor}

\usepackage{pgfplots}
\usetikzlibrary{pgfplots.statistics, pgfplots.colorbrewer, pgfplots.groupplots}
\usetikzlibrary{shapes.geometric, decorations.markings, arrows} 
\usepackage{pgfplotstable}
\pgfplotsset{compat=1.14}
\usepackage[binary-units=true]{siunitx}
\usepackage{algorithm, algorithmic}
\usepackage{todonotes}

\usepackage{multirow}
\usepackage{stfloats}
\usepackage[caption=false]{subfig}
\usepackage[final]{changes}

%% file: abstract.tex
\begin{abstract}
Learning-enabled components (LECs) are widely used in cyber-physical systems (CPS) since they can handle the uncertainty and variability of the environment and increase the level of autonomy. 
However, it has been shown that LECs such as deep neural networks (DNN) are not robust and adversarial examples can cause the model to make a false prediction. 
The paper considers the problem of efficiently detecting adversarial examples in LECs used for regression in CPS. 
The proposed approach is based on inductive conformal prediction and uses a regression model based on variational autoencoder. The architecture allows to take into consideration both the input and the neural network prediction for detecting adversarial, and more generally, out-of-distribution examples.
We demonstrate the method using an advanced emergency braking system implemented in an open source simulator for self-driving cars where a DNN is used to estimate the distance to an obstacle. The simulation results show
that the method can effectively detect adversarial examples with a short detection delay.

\end{abstract}

\begin{IEEEkeywords}
adversarial example detection, inductive conformal prediction, VAE based regression, self-driving vehicles
\end{IEEEkeywords}

%% file: introduction.tex
\section{Introduction}
\label{sec:intro}
%motivation
The use of learning-enabled components (LECs) such as deep neural networks (DNNs) is on the rise in many classes of cyber-physical systems (CPS).
Semi-autonomous and autonomous vehicles, in particular, are applications where LECs play a significant role.
%for perception, planning and control if they  are  complemented with  methods for analyzing and ensuring safety.  
Although DNNs can increase the level of the autonomy by handling the uncertainty and variability of the environment, multiple studies have shown that DNNs are not robust. For example, they are vulnerable to examples with small specially human-crafted perturbations in the input that cause false predictions~\cite{7780651, goodfellow2014explaining}.

%problem
LECs rely on design-time data-driven training to learn how to operate in unstructured and dynamic environments. 
%The LECs are successfully trained and evaluation of training and testing errors is satisfactory at design time.
However, LECs must be deployed and operate in a real system and the input data may be different than the data used for training and testing. Adversarial examples, in particular, can cause wrong predictions and impact system safety~\cite{sitawarin2018darts}. Adversarial example detection for CPS must have a low false alarm rate while being computational efficient for real-time monitoring.

%exsiting approach
\added{
Recently, there have been many efforts to defend against adversarial examples especially in the context of classification.
%Detection of the adversarial examples has received particular attention recently especially in the context of classification tasks. 
% ae
Autoencoder based methods, such as~\cite{DBLP:conf/ccs/MengC17, an2015variational},  utilize the reconstruction error or the reconstruction accuracy to detect adversarial examples. The autoencoders learn and encode the properties of the training data with latent representations. If a test example is from the same distribution as the training set, we expect a small reconstruction error.
%, whereas a large reconstruction error if the test examples is out of the training distribution or under attack.  
% svdd
The work in~\cite{pmlr-v80-ruff18a} presents an approach which aims to map the in-distribution input data into a hypersphere of minimum volume characterized by center $c$ and radius $R$. The adversarial examples are expected to be mapped to points out of the hypersphere.
Typically, existing methods do not take into consideration the dynamical behavior. 
}

In our previous work~\cite{CaiICCPS20}, we developed an approach which leverages inductive conformal prediction~\cite{balasubramanian2014conformal,volkhonskiy2017inductive} and inductive anomaly detection~\cite{laxhammar2015inductive} for out-of-distribution detection in learning-enabled CPS. The detection algorithm is based on a variational autoencoder (VAE) which is designed independently of the perception LEC, and the approach considers only the input data. Taking into account the predicted output can be beneficial for detection of adversarial examples especially since they are crafted so that a small perturbation of the inputs causes a large change in the output.
%our approach
%The proposed adversarial examples detection approach is based on the work~\cite{CaiICCPS20} which leverages the inductive conformal prediction~\cite{balasubramanian2014conformal,volkhonskiy2017inductive} and anomaly detection~\cite{laxhammar2015inductive} but uses variational autoencoders (VAEs) to learn models to efficiently compute the nonconformity of new inputs relative to the training set.
Instead of training the regression and VAE model used for detection independently, this paper uses a VAE-based regression model which integrates the regression model into the VAE and takes into account the predicted output for adversarial detection. 
The main contribution of the paper is an approach for detection of adversarial examples for regression LECs in CPS based on inductive conformal anomaly detection. In order to take into consideration the output of the regression LEC, the proposed approach employs a VAE-based regression model which is learned jointly combining the VAE and regressor by conditioning the latent representation of the VAE on the target variable of the regressor. The inductive conformal anomaly detection is based on the VAE-based regression model and can be performed efficiently for high-dimensional inputs. 

Another contribution is the empirical evaluation of the approach using an advanced emergency braking system (AEBS) implemented in CARLA~\cite{Dosovitskiy17}, an open source simulator for self-driving cars.
The AEBS uses a perception LEC to detect the nearest front obstacle and estimate the distance from the host vehicle based on the images captured by an onboard camera. 
The distance estimated by the LEC is used by a reinforcement learning controller together with the vehicle’s velocity to compute the desired braking in order to safely stop the vehicle. 
The fast gradient sign method  (FGSM)~\cite{goodfellow2014explaining} is used to generate adversarial examples for the perception LEC which cause large error in the distance estimation and result to a collision. It should be noted that although such attacks may be not physically realizable, they provide a framework to analyze the robustness of the LEC. The evaluation results show that the approach can detect such adversarial examples effectively with a short detection delay.

% organization
The rest of the paper is organized as follows. Section~\ref{sec:problem} introduces system model and formulates the problem. Section~\ref{sec:method} describes the VAE-based regression model and the detection algorithm based in inductive conformal anomaly detection. Section~\ref{sec:evaluation} presents the evaluation results using the AEBS. Section~\ref{sec:conclutions} concludes the paper.

%% file: problem.tex
\section{System Model and Problem Formulation}
\label{sec:problem}

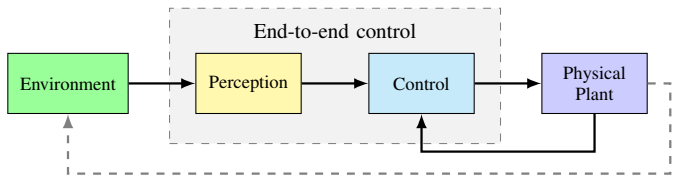
\begin{figure}[!ht]
	\centering
	\input{./figures/cps}
	\caption{Simplified CPS control architecture~\cite{CaiICCPS20}.}
	\label{fig:cps_arch}
\end{figure}

A simplified CPS architecture is shown in Fig.~\ref{fig:cps_arch}. The perception component observes and interprets the environment and feeds the information into the controller which, possibly using additional sensors (feedback from the plant), applies an action to the physical plant in order to achieve some task. In responding to the action, the state of the physical plant changes and the environment is observed and interpreted again to continue the system operation. LECs are extensively used for perception and control tasks in CPS in order to increase the level of autonomy.

Perception and control may be implemented using LECs (e.g., neural networks). An LEC can be designed using learning techniques such as supervised, unsupervised, and reinforcement learning. 
It is assumed that the LECs are successfully trained and evaluation of training and testing errors is satisfactory.  
However, adversarial examples with small input perturbations may cause the LECs to generate outputs with large errors and impact the safety of the system. 

The problem considered in this paper is detecting efficiently adversarial examples.
During the system operation, the inputs arrive one by one and an adversarial example can
cause large prediction error. 
The objective of the detection is to efficiently detect if the LEC input is adversarial. It should be noted that although such attacks may be not physically realizable, they provide a framework to analyze the robustness of the LEC.

%% file: figures/cps.tex
	\begin{tikzpicture}
	\node [draw, rectangle, minimum width=1.6cm, minimum height = 0.8cm,font=\scriptsize, fill=green!40](Env) at (0.8, 2.0) {Environment};
	\node [draw, dashed, rectangle, fill=gray!20, minimum width=4.4cm, minimum height = 1.8cm, font=\scriptsize, align=center, opacity=0.5] at (4.35,2.1)(e2e){};
	\node [font=\footnotesize] at (4.35, 2.70) {End-to-end control};
	\node [draw, rectangle, minimum width=1.4cm, minimum height = 0.8cm, font=\scriptsize, fill=yellow!40] at (3.2,2.0) (perception){Perception};
	\node [draw, rectangle, fill=cyan!20, minimum width=1.4cm, minimum height = 0.8cm, font=\scriptsize, align=center] at (5.5,2.0)(RL){Control};
	\node [draw, rectangle, fill=blue!20, minimum width=1.4cm, minimum height = 0.8cm, font=\scriptsize, align=center] at (7.8,2.0)(Vehicle){Physical \\ Plant};
	
	\draw [->, line width=0.3mm, >=latex](Env) -- (perception) node[pos=0.45, above, font=\scriptsize]{};
	\draw [->, line width=0.3mm, >=latex](perception)--(RL) node[midway, above, font=\scriptsize]{};
	\draw [->, line width=0.3mm, >=latex](RL)--(Vehicle) node[midway, above, font=\scriptsize]{};
	\draw [->, line width=0.3mm, >=latex](Vehicle.south)-- ++(0.0, -0.5) -| (RL.south) node[pos=0.25, above, font=\scriptsize]{};
	\draw [->,gray, line width=0.3mm, >=latex, dashed](Vehicle.east)-- ++(0.3, 0.0) -- ++(0.0, -1.2) -| (Env.south) node[pos=0.25, above, font=\scriptsize]{};
	%\draw[help lines](0,0) grid (8.5,3);
	\end{tikzpicture}

%% file: method.tex
\section{Variational Autoencoder for Regression and Adversarial Examples Detection}
\label{sec:method}
The detection algorithm proposed in this paper extends the work in~\cite{CaiICCPS20} by using a VAE-based regression model~\cite{zhao2019variational}.
The method is based on an LEC architecture which integrates the regression model into the VAE and uses inductive conformal anomaly detection~\cite{laxhammar2015inductive}.  

\subsection{Variational Autoencoder for Regression}
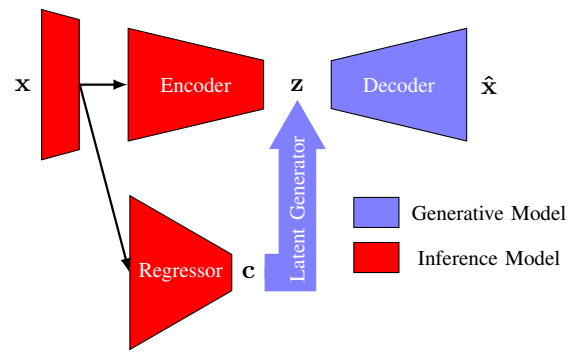
\begin{figure}[!ht]
	\centering
	\input{./figures/model.tex}
	\caption{VAE-based regression model\cite{zhao2019variational}.}
	\label{fig:model}
\end{figure}

A Variational Autoencoder (VAE) is a generative model whose encodings distribution is regularised during the training in order to ensure that its latent space has good properties allowing the generation of new data~\cite{DBLP:journals/corr/KingmaW13}. 
The objective is to model the relationship between the observation $x$ and the low-dimensional latent variable $z$. The architecture presented in~\cite{zhao2019variational} integrates a regression model into the generative model 
to disentangle the regression target variable from the latent space. The architecture of the VAE-based regression model is shown in Fig.~\ref{fig:model}. 

Similar to a traditional VAE, the encoder attempts to represent the input on the latent space using a Gaussian distribution $p(z)$. Instead of using a single Gaussian prior, the VAE-based regression model conditions the prior on the regression target variable $c$ such that the new prior $p(z|c)$, called \textit{latent generator}, can be used to sample a specific latent representation for a given target variable $c$. 
%On the other branch of the architecture, a regression network $q(c|x)$ with  additional output uncertainty of the prediction is used to inference the target variable. % XK: Are you sure about this sentence?
On the other branch of the architecture, a regression network $q(c|x)$ is used to inference the target variable and its uncertainty.
The loss function is defined as 
\begin{equation*} \label{eq:vaeLoss}
\begin{split}
\mathcal{L}(\theta, \phi_c, \phi_z; x) = &- D_{\text{KL}}(q_{\phi_c}(c|x) || p(c))\\
&+   \mathbb{E}_{z\sim q_{\phi_z}(z|x)}[\log p_\theta(x|z)]\\
&- \mathbb{E}_{c\sim q_{\phi_c}(c|x)}[ D_{\text{KL}}(q_{\phi_z}(z|x)||p(z|c))],
\end{split}
\end{equation*}
where $\theta$, $\phi_{z}$ and $\phi_{c}$ are the neural network parameters. The first term regularizes the prediction of $c$ with a ground-truth prior $p(c)$. The second term is the model fit trying to reconstruct the input from the latent representations. The third term is the KL divergence between the approximate posterior and the regression-specific prior $p(z|c)$. 
Both the VAE and the regression model can be jointly trained as a single network and the target variable is disentangled from the latent space.

\subsection{Adversarial Examples for Regression Model}
A common approach to generate adversarial examples is the fast gradient sign method (FGSM)~\cite{goodfellow2014explaining}, which computes an adversarial input by adding a small perturbation in the direction of the gradient. 
Instead of using this method to make the target network to misclassify the input, in our work, this FGSM is utilized to cause a large error in the output of the regression network. 

For a given input $x$ and a target output $y^{\text{target}}$, the neural network output is the regression estimation $c = f(x)$. The objective is to change $x$ to $\tilde{x}$ such that the error between $y^{\text{target}}$ and $f(\tilde{x})$ is minimized. This can be achieved by considering the cost function $J(\tilde{x}, y^{\text{target}}) = |f(\tilde{x}) - y^{\text{target}} |^2$ for generating the adversarial examples.

\subsection{Adversarial Detection}
\subsubsection{Conformal prediction}
\label{sec:cp}
Our method is based on inductive conformal prediction (ICP)~\cite{balasubramanian2014conformal}. The core idea of the method is to compute a \textit{nonconformity measure} defined by a function $A$ that assigns a value as \textit{nonconformity score} indicating how different a test example from the training data set. A large nonconformity score corresponds to a strange example with respect to the training data set. The original training set $\{ (x_1,y_1), \ldots, (x_l, y_l)\}$ is split into a proper training set $\{ (x_1,y_1), \ldots, (x_m, y_m)\}$ and a calibration set $\{ (x_{m+1},y_{m+1}), \ldots, (x_l, y_l)\}$. For each calibration example, the nonconformity scores $\alpha_{m+1}, \ldots, \alpha_{l}$ are precomputed relative to the proper training set. 
For a test example $(x^\prime_{k}, y^\prime_{k})$, the nonconformity score $\alpha^\prime_{k}$ is computed using the same way relative to the proper training set and the $p$-value is defined as the fraction of calibration examples that have nonconformity scores greater than or equal to the $\alpha^\prime_{k}$
\begin{equation}
p_k = \frac{|\{i=m+1, \ldots, l\} \, | \, \alpha_i \geq \alpha^\prime_k|}{l-m}.
\label{eq:p-value}
\end{equation}
If the $p$-value is smaller than a predefined threshold $\epsilon \in (0,1)$, the test example can be classified as a conformal anomaly. Using a single $p$-value for detecting the anomaly will make the detector oversensitive.
The robustness of the detector can be considerably improved if multiple $p$-values are used.
In \cite{vovk2005algorithmic}, it  is shown that if the test examples are independent and identically distributed (IID), the $p$-values are independent and uniformly distributed in $[0,1]$.
Further, a martingale test is used to test the hypothesis that the  $p$-values are independent and uniformly distributed,  and the martingale value is used as an indicator for the unusual test example~\cite{vovk2005algorithmic}. 
In~\cite{Fedorova:2012:PMT:3042573.3042693}, the \textit{simple mixture martingale} is used which is defined as 
\begin{equation}
M = \int_{0}^{1}\prod_{i=1}^{N}\epsilon p_i^{\epsilon-1}d\epsilon.
\label{eq:martingale}
\end{equation}
$ M $ will grow only if there are many small $p$-values in the sequence. More details about the martingales can be found in~\cite{Fedorova:2012:PMT:3042573.3042693}. 

The main idea in our approach is to use the generative VAE model to generate multiple examples sampling from the learned latent space probability distribution. The samples which are similar to the input are IID and can be used to generate $p$-values that are independent and uniformly distributed in $[0,1]$. Then, the martingale can be used to test this hypothesis and detect if the test example comes from the training data set distribution.

\subsubsection{Nonconformity measure and adversarial detection algorithm}
In~\cite{CaiICCPS20}, the reconstruction error (squared error between the input $x$ and the generated output $\hat{x}$) of the VAE model is used as the nonconformity measure
\begin{equation*}
\alpha = A_{\text{VAE}}(x, \hat{x}) = ||x-\hat{x}||^2.
\end{equation*}
In this paper, we use the same nonconformity measure based on the output of the VAE-based regression model which integrates the regression model and the VAE (Fig.~\ref{fig:model}). 

During the offline phase, 
the training set $\{ (x_1, y_1),  \ldots, (x_l, y_l)\}$ is split into a proper training set $\{ (x_1, y_1),  \ldots, (x_m, y_m)\}$ and a calibration set $\{ (x_{m+1}, y_{m+1}),  \ldots, (x_l, y_l)\}$. 
%In~\cite{CaiICCPS20}, two independent models are trained for regression and nonconformity measure separately. In this work, 
The VAE-based regression model is trained by utilizing the proper training data set.
%which combines the regression and the nonconformity measure in a single model. 
For each example in the calibration data set, the nonconformity score is precomputed using the 
nonconformity measure $A_{\text{VAE}}$.
 
At runtime, a sequence of test inputs $(x_1^\prime, \ldots, x_t^\prime, \ldots)$ is processed one-by-one. 
%and they are time-correlated, and therefore not independent. In order to satisfy the exchangeability assumption,  
Consider the input $x^\prime_t$. The method samples from the posterior distribution of the latent space to generate $N$ new examples $\hat{x}_{t,1}^\prime, \ldots, \hat{x}_{t,N}^\prime$, and
the nonconformity score $\alpha^\prime_{t,k}$ and the corresponding $p$-value for each generated example $\hat{x}_{t,k}^\prime$ are computed using $A_{\text{VAE}}$ and Eq.~(\ref{eq:p-value}).  
Since the generated examples are IID, the $p$-values $(p_{t,1}, \ldots, p_{t,N})$ are independent and uniformly distributed in $[0,1]$. Therefore, the martingale in Eq.~(\ref{eq:martingale}) can be used for detection.
The martingale denoted by $M_t$ has a large value if there are many small $p$-values in the sequence $(p_{t,1}, \ldots, p_{t,N})$ which indicates an out-of-distribution input. 

In order to robustly detect when the martingale becomes consistently large, \cite{CaiICCPS20} uses a stateful CUSUM detector to generate alarms for out-of-distribution inputs by keeping track of the historical information of the martingale values. 
The detector is defined as $S_1=0$ and $S_{t} = \max(0, S_{t-1} + M_{t-1} - \delta)$,  
where $\delta$ prevents $S_t$ from increasing consistently when the inputs are 
in the same distribution as the training data. An alarm is raised whenever $S_t$ is 
greater than a threshold $S_t > \tau $ which can be optimized using empirical
data~\cite{basseville1993detection}. Typically, after an alarm the test is 
reset with $ S_{t+1} = 0$.
Algorithm~\ref{alg:vae} describes the steps of the method. 

\begin{algorithm}[!ht]
	\caption{Adversarial examples detection using the VAE-based regression model}
	\label{alg:vae}
	\begin{algorithmic}[1]
		\renewcommand{\algorithmicrequire}{\textbf{Input:}}
		\renewcommand{\algorithmicensure}{\textbf{Output:}}
		\REQUIRE Input training set $\{ (x_1, y_1),  \ldots, (x_l, y_l)\}$, test examples $(x^\prime_1, \ldots, x^\prime_t, \ldots)$, number of calibration examples $l-m$, 
		number of examples to be sampled $N$, 
		stateful detector threshold $\tau$ and parameter $\delta$
		\ENSURE  Output boolean variable $Anom_{t}$, regression result $c_t$
		
		\renewcommand{\algorithmicrequire}{\textbf{Offline:}}
		\renewcommand{\algorithmicensure}{\textbf{Online:}}
		
		\REQUIRE 
		\STATE Split the training set $\{ (x_1, y_1),  \ldots, (x_l, y_l)\}$ into the proper training set $\{ (x_1, y_1),  \ldots, (x_m, y_m)\}$ and calibration set $\{ (x_{m+1}, y_{m+1}),  \ldots, (x_l, y_l)\}$
		\STATE Train a VAE-based regression model $f(\cdot)$ using the proper training set and get the corresponding nonconformity measure $A_{\text{VAE}}(\cdot)$
		\FOR {$j = m+1$ to $l$}
		\STATE Sample $\hat{x}_j$ using the trained VAE
		\STATE $ \alpha_j = A_{\text{VAE}}(x_j, \hat{x}_j) $
		\ENDFOR
		
		\ENSURE
		\FOR {$t = 1, 2, \ldots$}
		\FOR {$k = 1$ to $N$}
		\STATE Sample $ \hat{x}^\prime_{t,k} $ using the trained VAE
		\STATE $\alpha^\prime_{t,k} = A_{\text{VAE}}(x^\prime_t,\hat{x}^\prime_{t,k})$
		\STATE $p_{t,k} = \frac{|\{i=m+1,\ldots,l\} \,|\, \alpha_i \geq \alpha^\prime_{t,k}|}{l - m}$ 
		\ENDFOR
		\STATE $M_t =  \int_0^1 \prod_{k=1}^N \epsilon p_{t,k}^{\epsilon - 1} d\epsilon$
		\IF {$t=1$} 
		\STATE $S_t = 0$
		\ELSE
		\STATE $S_t = \max(0, S_{t-1} + M_{t-1} -\delta)$
		\ENDIF
		\STATE $Anom_{t} \leftarrow S_t > \tau $
		\STATE $c_t = f(x^\prime_t)$ 
		\ENDFOR
	\end{algorithmic} 
\end{algorithm}

%% file: figures/model.tex
\begin{tikzpicture}
\tikzstyle{vecArrow} = [line width=1.5mm, draw=blue!50,-triangle 60,postaction={draw, line width=5mm, shorten >=4.5mm, -}]

\node[draw,rotate=-90, trapezium, trapezium angle=60, trapezium stretches body, minimum height=0.5cm, minimum width=20mm, fill=red] at (1,2.5)(x){};

\node[draw,rotate=-90, trapezium, trapezium angle=70, trapezium stretches body, minimum height=1.8cm, minimum width=15mm, fill=red,font=\footnotesize, text=white] at (2.8,2.5)(encoder){\rotatebox{90}{Encoder}};

\node[draw,rotate=90, trapezium, trapezium angle=70, trapezium stretches body, minimum height=1.8cm, minimum width=15mm, fill=blue!50,font=\footnotesize, text=white] at (5.5,2.5)(decoder){\rotatebox{-90}{Decoder}};

\node[draw,rotate=-90, trapezium, trapezium angle=60, trapezium stretches body, minimum height=1.3cm, minimum width=15mm, fill=red, font=\footnotesize, text=white] at (2.6,0.0)(regressor){\rotatebox{90}{Regressor}};

\draw[->,line width=0.3mm, >=latex](x) -- (encoder);
\draw[->,line width=0.3mm, >=latex](x.north) -- (regressor.south);

\node [] at (0.5, 2.5) {$\mathbf{x}$};
\node [] at (4.15, 2.5) (z){$\mathbf{z}$};
\node [] at (6.7, 2.5) {${\mathbf{\hat{x}}}$};
\node [] at (3.5, 0.0) (c){$\mathbf{c}$};

\node[draw, rectangle, minimum width=6mm, minimum height=4mm, fill=blue!50] at (5.2,0.8)(){};
\node [font=\footnotesize, align=left] at (6.7,0.8) {Generative Model};

\node[draw, rectangle, minimum width=6mm, minimum height=4mm, fill=red] at (5.2,0.2)(){};
\node [font=\footnotesize,align=left] at (6.7,0.2) {Inference Model};

\draw[vecArrow](c)-|(z) node[text=white, font=\footnotesize] at (4.15, 0.85){\rotatebox{90}{Latent Generator}};
%\node [arrow box, 
%arrow box shaft width=0.6cm,
%inner sep=0.178cm, % should be half shaft width
%fill=blue!50, font=\footnotesize, text=white,
%arrow box arrows={north:0.5cm}] at (4.15,0.6)
%{\rotatebox{90}{Latent generator}};
%\draw[help lines](0,0) grid (8,4);

\end{tikzpicture}

%% file: evaluation.tex
\section{Evaluation}
\label{sec:evaluation}
We evaluate the adversarial detection method using an advanced emergency braking system (AEBS) implemented using CARLA\cite{Dosovitskiy17}. The experiments reported use CARLA 0.9.5 on a 16-core i7 desktop with $\SI{32}{\giga\byte}$ RAM and a single RTX 2080 GPU with $\SI{8}{\giga\byte}$ video memory.

\subsection{Experimental Setup}
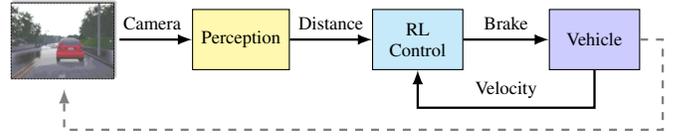
\begin{figure}[!ht]
	\centering
	\input{./figures/AEBS_architecture}
	\caption{Advanced emergency braking system architecture~\cite{CaiICCPS20}.}
	\label{fig:aebs_architecture}
\end{figure}

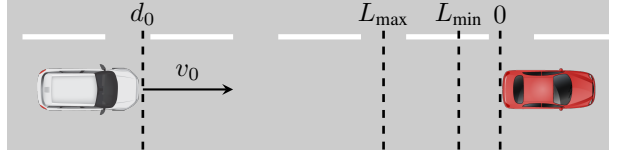
\begin{figure}[!ht]
	\centering
	\input{./figures/scenario}
	\caption{Illustration of advanced emergency braking system~\cite{CaiICCPS20}.}
	\label{fig:aebs_senario}
\end{figure}

A typical architecture (Fig.~\ref{fig:aebs_architecture}) and its desirable behavior (Fig.~\ref{fig:aebs_senario}) of the AEBS are presented in~\cite{CaiICCPS20}. 
The objective of the AEBS is to detect an obstacle and apply appropriate brake force in order to avoid a potential collision. The initial velocity of the host vehicle is $v_0$ and the initial distance between the host car and the obstacle is $d_0$. A perception LEC receives images captured by an on-board camera, detects the nearest front obstacle on the road, and estimates the distance. The estimated distance together with the velocity of the host vehicle are fed to a reinforcement learning controller whose objective is to generate the braking force for safely stopping the vehicle between $L_{\min}$ and $L_{\max}$. 

In order to simulate realistic scenarios, the initial velocity $v_0$ is uniformly sampled between $\SI[per-mode=symbol]{90}{\kilo\meter\per\hour}$ and $\SI[per-mode=symbol]{100}{\kilo\meter\per\hour}$, and the initial distance $d_0$ is based on the camera range in CARLA and is approximately $\SI{100}{\meter}$. 
%Besides, the precipitation parameter in the simulation is randomly sampled from the interval $[0,20]$ in the experiments. %XK: Are you using anything related to the precipitation parameter?
%FC: No.
The reinforcement learning controller is trained using the DDPG algorithm~\cite{DBLP:journals/corr/LillicrapHPHETS15} with $1000$ episodes and reward function which aims to stop the vehicle between $L_{\min}=\SI{1}{\meter}$ and $L_{\max}=\SI{3}{\meter}$.  It should be noted is that the sampling period used in the simulation is $\Delta t = \SI{1/20}{\second}$. The reinforcement learning controller is used only for simulation of the closed loop system and does not affect the proposed approach.
Details about the design of the reinforcement learning controller can be found in~\cite{DBLP:journals/tecs/TranCLMJK19}. % XK: Put the EMSOFT paper.

\subsection{LEC Training}
The data set for the perception and detector training consists of $8160$ images obtained by varying the initial distance $d_0$, initial velocity $v_0$ and precipitation. As described in \ref{sec:method}, the VAE and the regression network are jointly trained using the VAE-based regression model implemented as a convolutional neural network (CNN).

In the VAE-based regression model, the input firstly passes through four convolutional layers of $32/64/128/256 \times (5\times 5)$ filters with ELU activations and $2\times2$ max-pooling, and one fully connected layer of 1568 units with ELU activation. Then, the extracted features are fed into a fully connected layer with $1024$ units. The regressor shares the convolutional layers and also has two fully connected layers with $256/1$ units. 
After the regressor, two fully connected layers of $256/1024$ are used to yield distance-specific latent representations. The decoder has symmetric deconvolutional layers.  

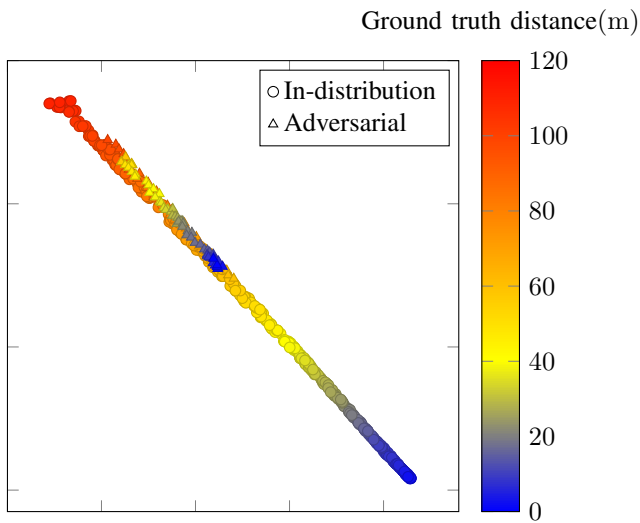
\begin{figure}[!ht]
	\centering
	\input{./figures/2dlatent_attack.tex}
	\caption{2D latent representations estimated by VAE-based regression model.}
	\label{fig:2dlatent}
\end{figure}

A simple two-phase learning schedule is employed with initial searching learning rate $\eta=10^{-4}$ for $250$ epochs, and subsequently fine-tuning $\eta=10^{-5}$ for $100$ epochs. After the training, the mean absolute error of the perception LEC for training and testing are $\SI{0.32}{\meter}$ and $\SI{0.40}{\meter}$ respectively. In addition, multi-dimensional scaling (MDS)~\cite{doi:10.1111/j.1745-3984.2003.tb01108.x} is implemented to seek a low-dimensional representation of the latent space of the VAE. Fig.~\ref{fig:2dlatent} shows the 2D representations of the latent space. % XK: Explain how you get the 2D representation - TNSE? % MDS, I have already explained it here. 
From the plots, see that the dimension related to the distance is disentangled from the latent space. 

A closed-loop simulation run is shown in Fig.~\ref{fig:inDistribution}. Initially, the distance between the host and the lead car is $\SI{98.03}{\meter}$, and the velocity of the host car is $\SI[per-mode=symbol]{96.95}{\kilo\meter\per\hour}$ ($=\SI[per-mode=symbol]{26.93}{\meter\per\second}$). 
After $140$ steps or $\SI{7.0}{\second}$, the host vehicle stops at 
$\SI{1.83}{\meter}$ from the lead car. 

\subsection{Adversarial Examples}
\begin{figure}[!ht]
	\centering
	\subfloat[Original image\label{fig:PhysicalAttack_a}]{
		\includegraphics[width=0.4\linewidth]{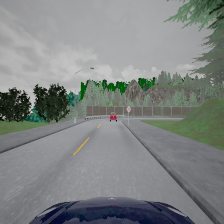}}
	\hfil
	\subfloat[Reconstructed original image\label{fig:PhysicalAttack_b}]{
		\includegraphics[width=0.4\linewidth]{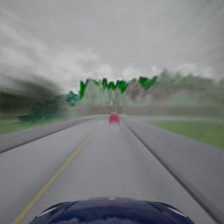}}
	\hfil
	\subfloat[Attacked image\label{fig:PhysicalAttack_c}]{
	\includegraphics[width=0.4\linewidth]{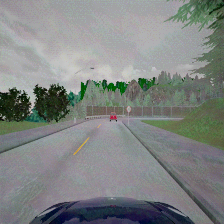}}
	\hfil
	\subfloat[Reconstructed attacked image\label{fig:PhysicalAttack_d}]{
	\includegraphics[width=0.4\linewidth]{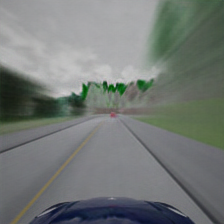}}
	\caption{Comparison of original image, image with attack and their reconstructed images.}
	\label{fig:PhysicalAttack}
\end{figure}

The FGSM is used to generate adversarial examples for the perception LEC. 
For a given input image $x_t$, the neural network output is the estimated distance $c_t = d_t = f(x_t)$. 
The objective is to modify $x_t$ by a small step $\epsilon=0.02$ in the direction that minimizes the loss to $\tilde{x}_t$ such that the predicted distance is close to the target value $y^{\text{target}}$. 
In our experiment, we set  $y^{\text{target}}=\SI{100}{\meter}$. 
The perception module predicts a large distance even when the car is very close to a stopped lead car. 
%In order to realize the real-time attack, we only apply the attack with a single step. 
%The constraint parameter of the input perturbation is set to $\epsilon=0.02$ and the learning rate $\alpha=0.1$. % XK: Not clear what are these. % FC: explain it in this paragraph.
%We set the 
A comparison of the original image and the adversarial image is shown in Fig.~\ref{fig:PhysicalAttack_a} and ~\ref{fig:PhysicalAttack_c}. 

\subsection{Simulation Results}
In order to analyze the performance of the detection method, we generate multiple simulation episodes which contain normal and adversarial examples. 
%For the adversarial episodes, we control the time step the attacks involved, which is uniformly sampled from $\{20, 21, \ldots, 60 \}$. % XK: Not clear what "control the time step ..." means. Is the following correct?
Adversarial examples are inserted in the simulation in a time step uniformly sampled from $\{20, 21, \ldots, 60 \}$.

To illustrate the approach, we compare two episodes and plot the ground truth and predicted distance to the lead stopped car, the velocity of the host car, the $p$-values and the $S$-value of the CUSUM detector (computed using the logarithm of $M_t$). We use $N=10$ for the number of the examples generated by the VAE and $\delta=12, \tau=80$ for the parameters of the CUSUM detector. Since $M_t$ becomes very large, we choose to use $\log M_t$ to show the results.

\begin{figure}[!ht]
	\centering
	\input{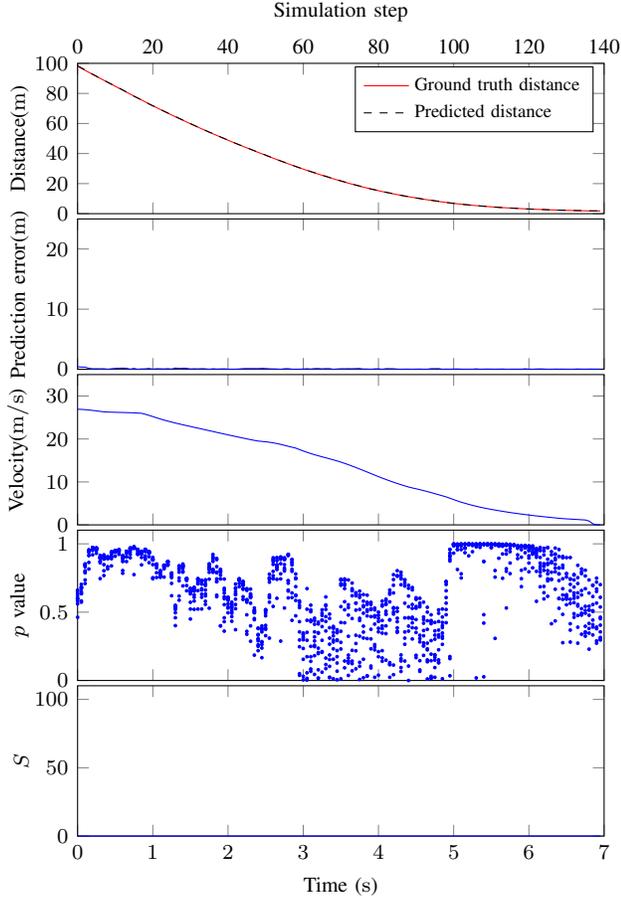}
	\caption{Episode with normal input examples.}
	\label{fig:inDistribution}
\end{figure}

\begin{figure}[!ht]
	\centering
	\input{./figures/attackPerception.tex}
	\caption{Episode with adversarial examples.}
	\label{fig:AttackPerception}
\end{figure}
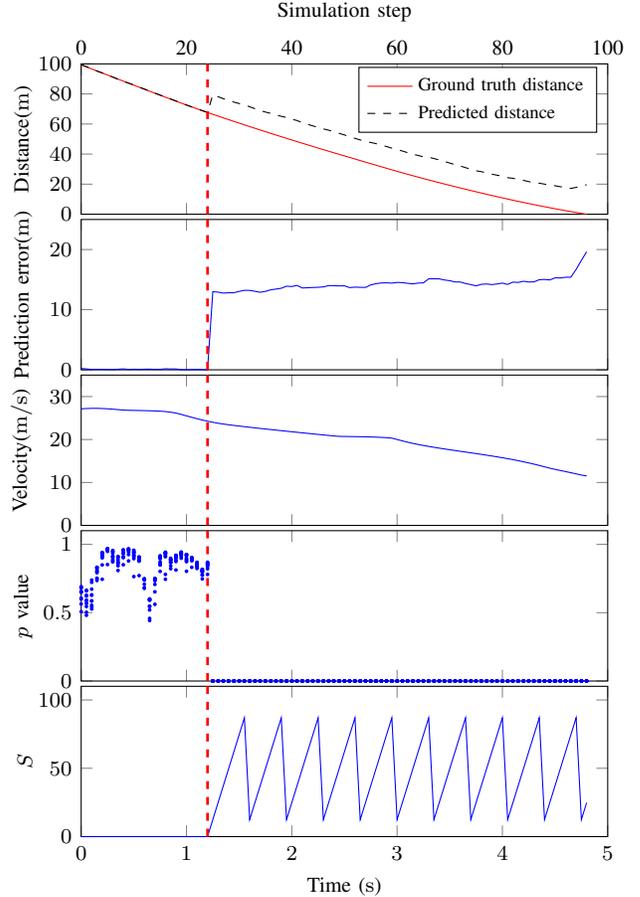

% in distribution
The normal case is shown in Fig~\ref{fig:inDistribution}. The $p$-values are randomly distributed between $0$ and $1$, and the martingale is small indicating the inputs are in distribution. 
% attack 
An episode with adversarial examples is shown in Fig~\ref{fig:AttackPerception}.  The adversarial example starts at $\SI{1.20}{\second}$ trying to cause the regression network to predict a larger distance than the actual one. The error starts to increase and reaches almost $\SI{20}{\meter}$. The controller is misled by the perception LEC and  does  not  stop  the  car  in time which  collides  with  the  lead  car (velocity is greater than $0$ when ground truth distance comes to $0$). The $p$-values become smaller and the detector indicates the test inputs are not from the same distribution as the training data set.  The delay for detection is smaller than $10$ frames or $\SI{0.5}{\second}$ (the sampling rate is $\SI{20}{\hertz}$).

We plot the 2D representations of the latent space generated by VAE using one episode of adversarial examples in Fig.~\ref{fig:2dlatent}. Since the latent space is conditioned by the regression prediction, the latent representations of the adversarial examples lie on regions of the space that correspond to different predictions of the regression network. While reconstructing the input for an adversarial example, the VAE generates the new examples from such a region. For example, an image reflecting a small distance is mapped to the region indicating a large distance for an adversarial example. 
Thus, the reconstructed image is sampled from the wrong region and reflects a large distance, and therefore, the reconstruction error or nonconformity score is large. We compare the original image, adversarial example, and their corresponding reconstructed images in Fig.~\ref{fig:PhysicalAttack}. % XK: make a pass in this paragraph! %FC: did it

We also evaluate the approach for $100$ normal episodes and $100$ adversarial episodes by considering different values of $N$.
We run each episode and if an alarm is raised, we stop the simulation, and we check if the alarm is false. We compute the detection delay as the number of frames from the adversarial example corresponding to the alarm raised. 
%We use a simple search to select the optimal detector parameters $\delta$  and $\tau$ for making false alarms less than $4$. % XK: What is 4? 
Table~\ref{tab:vae-based AEBS} shows the false alarms and average delay for different $N$, $\delta$ and $\tau$.  Since the $p$-values for the adversarial examples are almost $0$, the number of the false alarms is very small and the detection delay is smaller than $10$ frames or $\SI{0.5}{\second}$.

\begin{table}[!ht]
	\centering
	\caption{VAE-based detection.}
	\label{tab:vae-based AEBS}
	\begin{tabular}{c c c c}
		\hline
		\multicolumn{1}{m{0.18\columnwidth}}{\centering Parameters $(N, \delta, \tau)$} 
		& \multicolumn{1}{m{0.20\columnwidth}}{\centering False positive}
		& \multicolumn{1}{m{0.20\columnwidth}}{\centering False negative}    
		& \multicolumn{1}{m{0.21\columnwidth}}{\centering Average delay (frames)}  \\
		
		\hline
		$5, 6, 6$ & $0/100$ & $0/100$ & $1.0$\\
		\hline
		$5, 7, 23$ & $0/100$ & $0/100$ & $4.0$\\
		
		\hline
		$10, 10, 62$ & $0/100$ & $0/100$ & $4.0$\\
		\hline
		$10, 12, 80$ & $0/100$ & $0/100$ & $6.0$\\
		
		\hline
		$20, 18, 120$ & $0/100$ & $0/100$ & $3.0$\\
		\hline
		$20, 20, 280$ & $0/100$ & $0/100$ & $9.0$\\
		\hline
	\end{tabular}
\end{table}

\subsection{Computational Efficiency}
The VAE-based regression model can predict the target variable and compute the nonconformity score in real-time without storing training data.
Table~\ref{tab:real-time} reports the minimum $(\min)$, first quartile $(Q_1)$, second quartile or median $(Q_2)$, third quartile $(Q_3)$, and maximum $(\max)$ of (1) the execution times of the LECs in AEBS and (2) the execution times of the VAE-based detectors for different values of $N$.

Since the VAE-based detection uses $N$ examples in each time step, the execution time is proportional to the number of examples generated $N$. 
The execution times are much smaller than the sampling time ($\SI{50}{\milli\second}$ in AEBS, and thus, the methods can be used for real-time out-of-distribution detection.

\begin{table}[!ht]
	\centering
	\caption{Execution times.}
	\label{tab:real-time}
	\begin{tabular}{c c c c c c c}
		\hline
		\multicolumn{1}{m{0.05\columnwidth}}{} 
		& \multicolumn{1}{m{0.05\columnwidth}}{\centering $N$} 
		& \multicolumn{1}{m{0.08\columnwidth}}{\centering $\min$ (ms)}
		& \multicolumn{1}{m{0.08\columnwidth}}{\centering $Q_1$ (ms)}
		& \multicolumn{1}{m{0.08\columnwidth}}{\centering $Q_2$ (ms)}
		& \multicolumn{1}{m{0.08\columnwidth}}{\centering $Q_3$ (ms)}
		& \multicolumn{1}{m{0.08\columnwidth}}{\centering $\max$ (ms)} \\
		\hline
		AEBS &N/A & 3.49 & 3.86 & 3.93 & 3.98 & 4.22 \\
		\hline
		\multirow{3}{*}{VAE} &5 & 18.90 & 18.96 & 18.99 & 19.02 & 19.08 \\
		& 10 & 37.89 & 37.96 & 37.99 & 38.04 & 38.10 \\
		& 20 & 76.32 & 76.47 & 76.52 & 76.71 & 77.32 \\
		\hline
	\end{tabular}
\end{table}

%% file: figures/AEBS_architecture.tex
\begin{tikzpicture}
\node [](Env) at (0.75, 2.0) {\includegraphics[width=.08\textwidth]{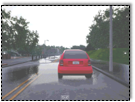}};
\node [draw, rectangle, minimum width=1.2cm, minimum height = 0.8cm, font=\scriptsize, fill=yellow!40] at (3.1,2.0) (perception){Perception};
%\node[font=\scriptsize, align=center] at (1.2, 2.7) {Environment};
\node [draw, rectangle, fill=cyan!20, minimum width=1.2cm, minimum height = 0.8cm, font=\scriptsize, align=center] at (5.45,2.0)(RL){RL \\ Control};
\node [draw, rectangle, fill=blue!20, minimum width=1.2cm, minimum height = 0.8cm, font=\scriptsize, align=center] at (7.8,2.0)(Vehicle){Vehicle};
\draw [->, line width=0.3mm, >=latex]($(Env.east)+(-0.1,0)$) -- ($(perception.west) + (-0.0, 0) $) node[pos=0.45, above, font=\scriptsize]{Camera};
\draw [->, line width=0.3mm, >=latex](perception)--(RL) node[midway, above, font=\scriptsize]{Distance};
\draw [->, line width=0.3mm, >=latex](RL)--(Vehicle) node[midway, above, font=\scriptsize]{Brake};
\draw [->, line width=0.3mm, >=latex](Vehicle.south)-- ++(0.0, -0.5) -| (RL.south) node[pos=0.25, above, font=\scriptsize]{Velocity};
\draw [->,gray, line width=0.3mm, >=latex, dashed](Vehicle.east)-- ++(0.3, 0.0) -- ++(0.0, -1.2) -| (Env.south) node[pos=0.25, above, font=\scriptsize]{};
%\draw[help lines](0,0) grid (9,3);

\end{tikzpicture}

%% file: figures/scenario.tex
\begin{tikzpicture}
\fill [gray!40] (0,0) rectangle (8,2);
\draw [white, line width=2pt] (0.2,1.5) -- (1.3, 1.5);
\draw [white, line width=2pt] (1.9,1.5) -- (3.0, 1.5);
\draw [white, line width=2pt] (3.6,1.5) -- (4.7, 1.5);
\draw [white, line width=2pt] (5.3,1.5) -- (6.4, 1.5);
\draw [white, line width=2pt] (7.0,1.5) -- (8.0, 1.5);
%\draw [white, line width=2pt] (8.7,1.5) -- (9.8, 1.5);
\draw [line width=1pt, dashed] (1.8,0.0) -- (1.8, 1.6);
\draw [line width=1pt, dashed] (6.55,0.0) -- (6.55, 1.6);
\draw [line width=1pt, dashed] (5.0,0.0) -- (5.0, 1.6);
\draw [line width=1pt, dashed] (6.0,0.0) -- (6.0, 1.6);
\node [] at (1.8, 1.8) {$d_0$};
\node [] at (6.0, 1.8) {$L_{\text{min}}$};
\node [] at (5.0, 1.8) {$L_{\text{max}}$};
\node [] at (6.55, 1.8) {$0$};
\node [rotate=180](vehicle) at (1.1, 0.8) {\includegraphics[width=.08\textwidth]{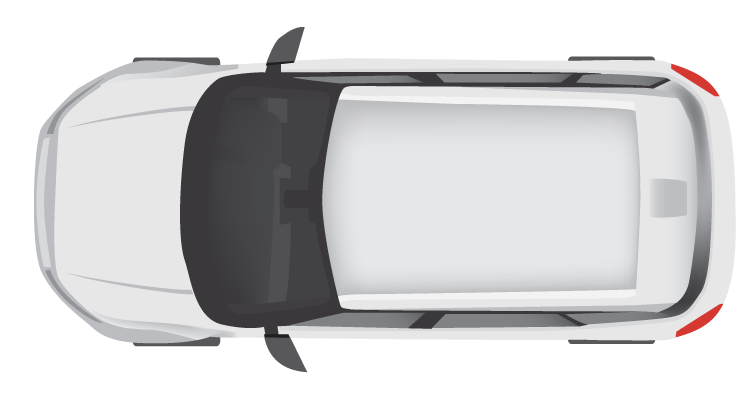}};
\node [rotate=180](leader_vehicle) at (7.2, 0.8) {\includegraphics[width=.07\textwidth]{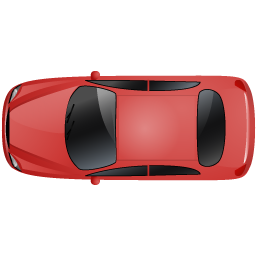}};
\draw [->, >=stealth, thick] (1.8, 0.8) -- (3.0, 0.8) node [midway, above]{$v_0$};
%\draw[help lines](0,0) grid (10,3);
\end{tikzpicture}

%% file: figures/2dlatent_attack.tex
% This file was created by matlab2tikz.
%
%The latest updates can be retrieved from
%  http://www.mathworks.com/matlabcentral/fileexchange/22022-matlab2tikz-matlab2tikz
%where you can also make suggestions and rate matlab2tikz.
%
\begin{tikzpicture}
\pgfplotsset{
	scale only axis,
	width=6.0cm,
	height=6.0cm,
	%major grid style={dotted, gray}
}

\begin{axis}[colorbar, colorbar style={title={Ground truth distance$(\SI{}{\meter})$}}, 
%visualization depends on=\thisrow{w2}\as\wtwo,
%scatter/@pre marker code/.code={
%	\ifnum\wtwo<2
%		\def\markopts{mark=square*}
%	\else
%		\def\markopts{mark=o}
%	\fi
%	\expandafter\scope\expandafter[\markopts]
%},
%scatter/@post marker code/.code={
%	\endscope
%},
legend cell align={left},
xmin = -15, xmax=9.0,
ymin = -23, ymax=40,
xticklabels={,,}, yticklabels={,,}
]
\addplot+[only marks,scatter, 
point meta=explicit,
point meta min={0}, point meta max={120},forget plot]
table[x=x,y=y,meta=w1]{
x y w1
3.393711590114047 -9.518587474565567 19.474879503250122
4.91056040342709 -13.961356497580947 9.500783085823059
-6.033282788443811 17.054402326598087 73.99285554885864
5.793323444004395 -16.50462438481461 6.064571142196655
4.733188598310694 -13.470897512274988 10.467279553413391
-8.91996223254177 25.613242405868174 91.48200988769531
5.746849310634828 -16.36242376671571 6.558487415313721
-10.16479190691974 28.150173839564346 96.84176445007324
-2.9038302176486432 8.271115539217497 55.95680236816406
-6.405606686051502 18.484060887665983 76.90454006195068
4.912358766228737 -13.872610969247129 9.638073742389679
-10.68830892577663 30.251300806826894 100.95498561859131
-7.877410259987387 22.138143951744464 84.49368953704834
5.1092276503063845 -14.911604423922542 8.505679070949554
4.58297166186866 -13.167632958437219 11.352345943450928
0.24048874122902397 -0.4557079424525485 37.956833839416504
-9.272343210859976 26.141274841497435 92.65647411346436
-1.0078425283515287 2.778571598485243 44.86932992935181
4.8292629174608965 -13.725071140377569 9.92857038974762
-4.738362450756313 13.915984996565095 67.61401176452637
6.147074067854114 -17.694211955167024 3.1746697425842285
4.872909339832974 -13.95891047010498 9.524948000907898
-2.9012694267424286 8.633662804106823 56.56761646270752
3.3842255358356126 -9.81034235963876 19.050318002700806
5.628893442878187 -16.234008828510166 6.839887797832489
4.276096409676799 -12.229732214568548 14.051571786403656
-3.503161023444125 10.636233408703454 60.675315856933594
6.396905554035426 -18.374271953349954 1.9166064262390137
4.601054545583374 -13.193032770742878 11.244986057281494
6.358547234238065 -18.209858444062114 2.221456468105316
4.211640931876501 -12.213892850435743 14.132204353809357
-0.058844811942559375 0.7801073683166305 40.41496753692627
5.838394510222307 -16.77069025515758 5.335111767053604
-9.765230405902452 28.256594156101414 96.77384376525879
3.3808930654938676 -9.453364123429758 19.460189938545227
6.419646777095757 -18.23898741454248 2.1054822206497192
6.074759625695656 -17.34202685434855 3.8611355423927307
-4.082458285231394 12.311036171641158 64.05532121658325
-1.5340522060614277 4.128744191385246 47.575507164001465
4.689731778932638 -13.560156722066196 10.38234829902649
-0.9191822757416322 2.489429892798087 44.38847780227661
-10.0719458835705 28.15003002081825 96.77624702453613
-4.071986675684043 11.123110585578699 61.89840316772461
-7.870652536886978 21.68081939070008 83.60335350036621
4.837181205866703 -13.708259839850896 9.911393523216248
-4.713420342471477 13.393078680276373 66.52899742126465
-2.4867100436936576 6.989835090190774 53.23503613471985
-2.8491182663636847 7.801576628152129 54.96477127075195
5.939781447793231 -17.00566832840669 4.548528045415878
-8.377105853077632 24.19286287372643 88.57632637023926
-1.4013566905805923 4.685512158421717 48.42944025993347
4.16244280100623 -11.753459301180218 15.078452825546265
-12.160565178083912 33.4579970644929 107.6693344116211
6.105089617732594 -17.393985685246182 3.808319717645645
5.822456102054081 -16.708487134222 5.5025191605091095
1.0401269655146326 -2.5858914685404466 33.327287435531616
0.8349598826141342 -1.7489040153739817 35.12447476387024
-11.419951839114685 33.03260770707594 106.4297604560852
-11.43750540113138 32.9766241714309 106.33918762207031
-3.7387590452758963 11.131435847581344 61.70428991317749
-0.4756869583973982 1.8746795337131943 42.8400993347168
4.661125641612363 -13.249603153139436 11.125566959381104
2.0538651535935437 -5.641957393422222 26.92587375640869
-11.136476718633036 30.866585079468166 102.35022783279419
5.429408435469354 -15.470003517997023 8.060736358165741
6.041474728489408 -17.257476715118603 4.049674719572067
-9.418263023951646 26.09524918324455 92.6676607131958
3.280183371851221 -9.174894796248294 20.067105889320374
0.5412567175862276 -1.0730805705026345 36.6889214515686
6.079302490351931 -17.35422500073724 3.9707350730895996
5.859181442031051 -16.724370381903082 5.582254976034164
-10.464269665366883 29.564110695091998 99.57635879516602
0.5978866306123118 -1.1079831618621394 36.5874981880188
0.620220232605804 -1.5093553716328727 35.765347480773926
1.0836322827427987 -2.6081028615957185 33.31496715545654
3.6264841304442164 -10.429106594921514 17.757643461227417
-4.496002930059479 12.730716996018932 65.21333456039429
-4.125442066077685 11.750938548813902 63.03339958190918
-7.729499476522346 22.430790259593596 84.94476556777954
-9.25126603448062 25.4040206776823 91.31844520568848
5.531106602998107 -15.691624351564856 7.860061526298523
5.017457600650392 -14.135332470570782 9.22903597354889
6.166290187987393 -17.510742759798678 3.357143774628639
5.723907250596073 -16.47544311170127 6.233103275299072
0.4304504097810415 -0.5711323214553194 37.660932540893555
2.9438887514237946 -8.327716700555863 21.465200185775757
5.652507849056987 -16.285784087234 6.749778091907501
-7.396719907745754 21.618867526953977 83.13740015029907
6.046784087360761 -17.257943564873653 4.089187681674957
2.8519996352938835 -7.928416573342249 22.336846590042114
5.972100267419584 -17.14106633740092 4.2378222942352295
-0.2588177981779973 0.9986201164597381 40.95053672790527
3.789077997519283 -10.558068737547645 17.439240217208862
2.218139022968335 -5.966357652594908 26.351956129074097
2.710870979985604 -7.390388832703117 24.031571745872498
4.549378765765201 -13.145805833551425 11.435099244117737
5.630165219216108 -16.245691630683993 6.789042055606842
-2.1127499303845445 6.436738570459234 52.059820890426636
-9.905768145645304 27.4456233372279 95.4048228263855
-9.866163423097289 27.443879926069492 95.37975311279297
2.7563377381318768 -7.7055386072893555 23.253800868988037
5.873778221140851 -16.83100458700574 5.113056153059006
1.3323484678931987 -3.840485478391796 30.527071952819824
6.013942240772669 -17.16901704689377 4.395101219415665
-6.371662799403421 18.233025900998918 76.39989852905273
-1.6356273564478296 5.308549536434366 49.67465043067932
6.179978352906463 -17.645604802058283 3.2030555605888367
-8.655237741047909 24.058171604171804 88.4952163696289
-9.913860843743981 27.49862739084916 95.50525903701782
-8.763390979829289 24.80048112959074 89.91717338562012
-7.812154449529728 22.420278024683643 84.88816738128662
6.283699812960875 -18.086869492361224 2.4385980516672134
4.606103857514422 -13.29667576201278 10.969039499759674
4.599514804207651 -13.185572135456596 11.327061653137207
-1.5002268653831516 4.799111689729147 48.69280815124512
-4.295264985425921 11.878465159709735 63.37629318237305
-4.939036941408002 13.542622293074182 66.93737983703613
0.5724428658541376 -1.5175753634846707 35.790313482284546
-5.6374245199657915 15.696769943538325 71.2854552268982
-8.48651240857683 23.591430045820665 87.54379749298096
5.832805207187734 -16.674196541555066 5.6524403393268585
4.47440210975211 -12.862622957782996 12.282115817070007
1.9984011480636155 -5.197712929322887 27.55783259868622
2.592086502376648 -7.235640507232703 24.531465768814087
6.044144914843344 -17.308729058635993 3.9885883033275604
2.781298043684324 -7.702425760590495 23.246312141418457
4.16376654573585 -11.760728386203922 15.067394971847534
2.4544381412727847 -6.568041698111136 25.587079524993896
5.119821829109223 -14.523981479678556 8.791642785072327
-2.2462133178802266 6.886421932101368 52.89993166923523
-9.273940156723832 25.495295989882646 91.49602174758911
-1.6091144203182632 4.682124636099112 48.54221820831299
-1.8889503378563017 5.737083242057542 50.61063051223755
-2.2481902518281647 6.277215418892364 51.80624842643738
-10.357001488324068 28.81757838817981 98.16390037536621
-5.69395205080764 16.448089251003495 72.73067235946655
-2.1759361406179263 6.2611993970470285 51.74595594406128
2.2832156634487335 -6.304164583179656 26.02056384086609
5.724914036886403 -16.30951418038287 6.5919095277786255
2.8368074704874573 -7.76398815519563 22.991187572479248
1.6896279477525347 -4.531822334246416 28.821802139282227
-5.497217672825037 14.99814974581515 70.0197172164917
3.9788731901879437 -11.503479336700355 15.693773031234741
0.26316036138523374 -0.08671590679221068 38.63667368888855
0.5115492891302709 -1.05206646909778 36.732598543167114
1.1607969245614336 -3.261476890917186 31.90568447113037
0.05803355619299701 0.5506608205815258 39.9288547039032
5.225194406802594 -15.053303709068985 8.410922884941101
-10.035442482980478 28.16298727710644 96.77693367004395
0.8905533115160083 -1.9509287547731398 34.71158266067505
5.991238726629318 -17.090141205679114 4.413732290267944
5.13128289893316 -14.884812429811236 8.551474213600159
5.666085626636941 -16.302622250165506 6.712930351495743
-10.401789056982249 28.77692376222477 98.11991930007935
-2.148395890359772 6.772325100962767 52.69378423690796
-4.151454387992146 12.49276310649314 64.6237564086914
5.891906062166289 -16.900455867220288 4.926247447729111
1.6448918877048975 -4.2901031633364735 29.332401752471924
4.953352356406123 -14.24977653244622 9.156872034072876
3.7650749271403456 -10.476045466280537 17.69452214241028
3.67240790871943 -10.591571835356488 17.45630979537964
0.5592216151907938 -1.5268589873400744 35.781118869781494
3.4564973623229487 -9.932021312525388 18.800357580184937
6.005425686159596 -17.132789481095383 4.33140903711319
-0.9598384011607209 2.5872008329705065 44.50499653816223
-8.677254132664595 24.183034612949847 88.74705076217651
-7.924606968573982 22.27712864651889 84.76651668548584
4.6631237364017855 -13.25066447628876 11.138302981853485
-1.5390173463041315 4.2413562834531255 47.65591263771057
0.9703085689785893 -2.391863138037265 33.768324851989746
5.751228836027445 -16.509755007188005 6.226454526185989
3.4351695262388846 -9.851978559837072 18.83959650993347
-11.36883207736178 32.98851836132959 106.31840944290161
-3.813885959525258 10.48685040005745 60.56217670440674
0.47245501190556644 -0.9013254924350197 37.02836036682129
1.726681549302408 -4.636825659597314 28.608555793762207
-7.416733669157611 20.824514211289674 81.72981977462769
-4.8681647773418435 14.033486554235358 67.84663438796997
0.8615614142112433 -2.4414230755515463 33.753401041030884
5.6149104076536025 -15.984219839462678 7.357490658760071
-1.2498189896093121 3.659211768107259 46.52258634567261
-2.357217777285649 7.040798617083232 53.24102997779846
0.9840311376153495 -2.1635352416909597 34.17550206184387
1.9565723338461873 -5.089722250617137 27.753514051437378
4.482800365704978 -12.706301587762479 12.716036438941956
3.103637668812089 -8.590881815712034 21.067965030670166
1.3992724271661525 -3.51965691606529 31.187353134155273
0.5719626571649836 -1.5983991946360754 35.63405156135559
-11.510458648558762 32.959304915832945 106.35347843170166
-1.7370039679331712 5.176918685986963 49.603031873703
-4.3021807137464245 12.825235296311956 65.30781269073486
5.82003962840784 -16.58443476573715 5.811566412448883
-1.0067412416330097 3.25496142177483 45.724854469299316
4.665534210315925 -13.358960657093975 10.743336081504822
3.4687615263877993 -9.745508509753243 18.969980478286743
-6.201005036283562 17.187212125363367 74.38173294067383
6.213334969378031 -17.882306036270602 2.824222147464752
5.0114038513344745 -14.62591190533889 8.798614740371704
1.7584086387689728 -4.926674607326386 28.07566523551941
-0.30382050233666175 0.8150589142372856 40.612138509750366
5.400274695148952 -15.224138032716649 8.190649151802063
-0.035628616031114335 0.28647525595278744 39.50116753578186
4.8264528858220075 -13.704677385405859 9.96400237083435
-3.109922968058558 8.532301731882718 56.54316544532776
-6.396012443744814 17.487152640975687 75.05609035491943
4.330613577750268 -12.56359296476753 13.215140104293823
5.5024847921914715 -15.929527244401575 7.52720832824707
-2.9862057835322204 8.275922868836412 55.97801685333252
6.06116944668458 -17.340700962702876 3.8559727370738983
6.042196803165176 -17.305403664649646 3.9656831324100494
-7.513522654181045 21.69809819597267 83.3849287033081
1.2749809925635562 -3.559429559090101 31.133240461349487
6.438852550061572 -18.345592083920266 1.9272925704717636
-9.281798696388893 26.83328368608304 93.9104175567627
6.00812375905112 -17.16841673072755 4.256514608860016
-9.34975951585209 26.156921362415368 92.7358603477478
-3.6783989896160234 10.486687252158044 60.50490617752075
-4.333155574440592 13.138012294407838 65.79941511154175
1.9122773875503853 -5.161547353780641 27.621143460273743
5.745270091183351 -16.408648464468463 6.306498348712921
-8.088973974664231 22.863293042448333 85.97898960113525
-6.008040623717713 17.679881051617432 75.13407468795776
0.8638166345161792 -2.38401662336554 33.869179487228394
-6.430921968810519 18.544557373047564 76.99855327606201
-8.270736657105932 23.646024714948773 87.49977350234985
-11.194232090297747 31.509145240796073 103.54144334793091
6.037663773216342 -17.2799074441016 4.022270143032074
4.076784553942337 -11.526737166275133 15.585184693336487
-7.894889615528731 23.090979164650417 86.27366781234741
6.308244767706136 -17.862814055256667 2.7144628018140793
-4.038459502310175 11.258052152642035 62.127907276153564
4.726570626394311 -13.586891055757478 10.291281044483185
-12.721437186027517 34.04228908714173 109.09594058990479
5.589455720018649 -16.078947161890852 7.199661433696747
3.636403963096853 -10.177796456474105 18.16285729408264
0.20089474471918967 -0.16116898380738992 38.655123710632324
4.656826318213788 -13.240745468857495 11.119774281978607
5.785877125937773 -16.481860800548887 6.2393344938755035
-1.506364953129557 4.674263410611579 48.41582536697388
0.22119731992325908 0.06407034565576254 38.958234786987305
5.814557069074819 -16.700744560903182 5.513775050640106
-0.7618402566343694 2.0745235649233176 43.361817598342896
-6.316499562101102 17.687731846425695 75.40811777114868
1.0852559157978714 -2.7742684736744354 32.97722339630127
-9.116825103859538 26.21498695137229 92.69222974777222
5.677333325103126 -16.367223852536007 6.4918094873428345
-6.020925750532172 17.63027989771357 75.06699800491333
0.17088532263011422 -0.20184334951281238 38.47387075424194
6.0696503540784725 -17.36038912395362 3.825947642326355
5.681889093196228 -16.373379934521708 6.4545124769210815
-11.386894400399893 31.462273322269546 103.58234167098999
-4.863054609827844 13.994870571124144 67.76918649673462
1.0934062134136755 -2.809499681106332 32.85032272338867
-4.592804215117672 13.727061383616547 67.05292224884033
5.147423601202448 -15.00871532875518 8.424670100212097
1.014258897584641 -2.8637546010180848 32.815550565719604
6.362163693968413 -18.040958031298064 2.4451401829719543
-6.790640183490896 19.630280060490044 79.28420305252075
3.734269214791161 -10.826683469473315 17.039299607276917
1.5265931344706318 -3.9219984560760195 30.207170248031616
-2.419405128925042 6.402177376777055 52.15724229812622
-7.907725218067002 22.237897160556834 84.66901302337646
-9.900114631880498 27.510695071250943 95.51725387573242
5.960738639210243 -16.985625085135617 4.80779305100441
1.8267600421457901 -5.206211282924997 27.63392686843872
-7.466254578422649 20.792311180564038 81.78999423980713
3.918141300594298 -11.159969834532648 16.28512144088745
-5.93956380808554 16.31793688668117 72.59284257888794
-1.1959301786636092 3.7164671989314715 46.57276153564453
5.741600632722224 -16.350675699925183 6.460628807544708
0.8047721702431373 -2.2027174558077505 34.219619035720825
0.6924403178495375 -1.489787676317979 35.75772285461426
-7.680246901240914 22.450735824554332 84.89044189453125
3.84026859996266 -10.750258451376018 17.081483602523804
4.56701447848895 -13.145258975393222 11.386089026927948
1.9628677273093267 -5.13715046919912 27.69153892993927
0.09976146252522547 0.011043248928573057 38.90750169754028
5.044200924278251 -14.452600850417884 8.902992904186249
-3.572542903182854 10.005815277202325 59.54192876815796
-10.83510239884597 30.871698029974635 102.16555595397949
-9.761036871750518 26.74081039286074 94.04947757720947
-5.197656887756206 15.112802487058802 70.08702278137207
-2.3923754950750578 6.993519630968231 53.24740290641785
2.7536326487107936 -7.540052966175764 23.67833912372589
4.0611691800613805 -11.42676367822593 15.772300958633423
-4.998268044070893 14.83255906997071 69.38223123550415
4.599201971477016 -13.347019995111422 10.796651244163513
5.23539733237257 -14.780782214201622 8.564389050006866
-7.400612637373118 20.919161468546207 81.8690299987793
2.276388069334956 -6.1513445873047905 26.128225922584534
2.377536277403977 -6.799988617769787 25.278815031051636
-11.179874069452715 30.828527699560507 102.31025218963623
4.619405191139458 -13.226068459965873 11.14759773015976
6.174105814641613 -17.762375999807226 3.0469688773155212
-4.34487285317004 12.391362205356005 64.52468633651733
-0.010423228999860234 0.545568587739063 39.948638677597046
-4.859493004586367 14.270828963820659 68.2487440109253
-6.096245760257788 17.29088658314829 74.46397304534912
-3.531852520325263 9.929965706910815 59.40738916397095
-8.806990109806279 24.774729840695983 89.89628791809082
3.7091106861574703 -10.773494474402835 17.124788761138916
1.2370376149107125 -3.128904522651025 32.09752321243286
4.892516541384025 -13.939478974526427 9.561392068862915
-5.808492423460604 15.971840356329803 71.9178557395935
2.917858112266244 -8.168526883438533 21.772974729537964
-10.194056504480484 28.14821489500075 96.85210704803467
-9.5459091626523 27.586504902508484 95.42896270751953
4.04043590672557 -11.75855167008484 15.15346348285675
0.5566629025058012 -1.1025004556445084 36.560014486312866
4.414969346781407 -12.60888293892934 12.988806366920471
4.948626080161677 -14.291540048660643 9.089927673339844
-0.9568740207745211 3.0484239101974366 45.343955755233765
-5.76393552500638 16.653748112656697 73.0967903137207
2.76126950794214 -7.486705817666854 23.67772400379181
5.343245665554893 -15.433838777626447 8.048963248729706
4.6449830580848 -13.502740347878241 10.494140982627869
2.4389823043838765 -6.820632630986843 25.29526948928833
-0.698973910867466 2.3534805470702085 43.86348009109497
5.798044249082333 -16.64073125271194 5.762253105640411
5.969020855965386 -17.02047092158351 4.599941074848175
5.036402598422631 -14.18590074014854 9.182502329349518
-3.5085769919436243 10.58630292836642 60.58084487915039
-12.368285384237074 33.52546680895619 107.9305100440979
2.2762667382841397 -6.002698917530221 26.387507915496826
-9.33175882850429 26.945535103679145 94.14452075958252
-4.06757496513658 11.794809762063005 63.12073230743408
5.196557394247091 -15.114438926931605 8.331405222415924
0.7683272305296133 -1.5172907973854939 35.66362738609314
-5.986373894377669 16.360009890487042 72.76287317276001
4.660181118997073 -13.293228769904323 10.926277041435242
6.459242701834502 -18.357829880187097 1.9509340077638626
4.77987084585608 -13.885114248368886 9.686456322669983
4.4844196300619785 -12.907549996262608 12.201110422611237
-6.2467372133499675 18.316796403344807 76.41614198684692
-9.158070659468457 26.24434592146112 92.7697491645813
6.042363598051863 -17.292008555519452 3.9818675816059113
4.933231519800194 -14.162240878080192 9.24176573753357
-9.827598390390339 28.224713299630473 96.75546169281006
5.590984161963251 -15.916717901761638 7.493247985839844
6.059834014247484 -17.345579740558122 3.8568006455898285
-10.675028818423803 29.444579762033076 99.49672222137451
1.0351756091470077 -2.831415180934194 32.866222858428955
2.88895180255246 -8.27415857974088 21.76216721534729
4.627150722985281 -13.179233180225543 11.253191828727722
5.698811686940147 -16.29884855403313 6.703577041625977
4.90550977680961 -13.994349450300842 9.495822787284851
1.7789860625952536 -4.727858211483157 28.467326760292053
3.918288381314531 -11.194972921833847 16.32802963256836
-8.033890234692628 22.272370854162826 84.7893476486206
4.02579981080025 -11.406935249677499 15.75472891330719
4.629303425291811 -13.37987970682935 10.753475725650787
-9.22199319118287 25.404200734631214 91.30005598068237
5.971980735553546 -17.11827991203503 4.36885803937912
-8.510984956860712 23.41461888272809 87.2622013092041
-3.7685924768061105 10.77207285003348 61.08640193939209
-1.5055952740470746 4.235040349976639 47.61307239532471
1.1391207596534318 -2.7003960354256424 33.028128147125244
-9.016551971974438 24.80949186822705 90.0955867767334
4.011657250559126 -11.516860640634873 15.648559927940369
6.026317367394726 -17.253732061148238 3.9335569739341736
1.9048765385300002 -4.942567543063049 27.986315488815308
0.6301200281344601 -1.5794291355861692 35.61901330947876
4.904243540324726 -13.968546067224992 9.546952843666077
-4.595941303947901 13.360420176484025 66.3991928100586
-4.748697565841921 14.024362519850293 67.72901058197021
-3.5971265876290643 9.872922426472849 59.34812664985657
5.281219209668142 -15.30112511185999 8.184565007686615
0.21091896695937437 -0.2041964048529699 38.50525259971619
6.287671925207854 -18.098995201088123 2.421894147992134
0.8493203093523451 -2.3994484330567807 33.88828754425049
-10.996885163514017 30.847348818018617 102.22499370574951
-10.133980066104884 28.10521096032457 96.7347264289856
-12.77670443551062 34.00287830357482 109.06465530395508
-2.0549689544100294 6.220716215452188 51.581504344940186
3.218649189934877 -8.777754635711108 20.715726613998413
-8.749053157353003 24.231552249839783 88.88044595718384
-8.465554674512989 24.268474309628 88.7652325630188
-9.550124041758949 26.829726241611954 94.07429695129395
6.4032160084988945 -18.366593032950057 1.8517573177814484
-1.7410595715887507 4.837418904992851 48.92728328704834
0.29722572598701236 -0.18670015525817907 38.47730755805969
2.597594211415441 -7.287404912112072 24.441906809806824
4.772283413658821 -13.534937068789455 10.342583656311035
1.0305032398353755 -2.3071671828174223 33.89938473701477
-11.731787854972488 34.331117578243635 108.97284507751465
-6.648291505835975 19.60489787829161 79.06287431716919
5.5506164791188635 -16.029807702597285 7.316198945045471
4.407386711233199 -12.534269958922032 13.125302195549011
-10.60228638202083 29.48659108866488 99.52474594116211
-1.7569296844371671 5.461291600382868 50.01483678817749
-0.25323320012581985 0.7411897363795971 40.46664118766785
5.997134058456716 -17.098408512341134 4.527750313282013
-10.391260059131767 28.771913900690286 98.10380458831787
6.190805274172445 -17.77876987223089 2.904404103755951
5.892760762138693 -16.879542372053084 4.94431346654892
5.507731281450445 -15.804870581521964 7.738924026489258
-2.3226994700773496 6.165261125836383 51.65379881858826
-9.624688317265031 27.590079720363725 95.4875922203064
-5.7395172628784845 15.775292547587046 71.4870285987854
1.0203892010582827 -2.7941635781755334 32.96909809112549
1.3534427302338736 -3.808008666209718 30.54630160331726
-2.6388560224252826 7.232987943182513 53.85822057723999
-3.345133492549251 9.977597477944812 59.418861865997314
-11.631573494855918 34.36509477030938 108.9598274230957
-1.849949865538723 5.729019429555445 50.5671501159668
5.572457225440512 -16.098702342671178 7.177750915288925
-10.483215952509044 29.479365219387716 99.43465232849121
3.6651359557845096 -10.612883309471783 17.464200854301453
3.1462026880879437 -9.062588484102761 20.341025590896606
-0.6357840878593692 2.282765906159324 43.767303228378296
-5.481087238841072 14.998147692900003 70.06457090377808
6.08651572782972 -17.42150517578818 3.848842531442642
-11.722292048298717 32.888720412029286 106.36080265045166
-11.720360954037115 32.831274263837685 106.25616073608398
-5.892696312531384 17.35178010895841 74.4820761680603
2.8039755970280567 -7.806863596268539 22.95387625694275
-9.138080011409231 26.198940431608357 92.67509937286377
2.0674318227773765 -5.323313831511956 27.339200377464294
6.3354209644632755 -18.052871478706212 2.4687693268060684
5.870824250445846 -16.853981080098713 5.114404410123825
0.011593409207489493 -0.06883069947915288 38.84545683860779
2.3182288799224064 -6.601579593389933 25.57369351387024
-8.316015983855554 23.50557001539483 87.2834587097168
6.233956455531257 -17.706257162626216 3.0086667090654373
2.4423399418209417 -6.888874546405663 25.22914409637451
5.608026555432587 -15.94730156540308 7.4287645518779755
-10.08441293199527 28.16643818143329 96.81617259979248
6.4372682362274904 -18.299585398486894 1.9948115944862366
-12.237859047352341 33.49550116505632 107.78855323791504
-8.963766807200487 24.746005566617992 89.94619846343994
-7.71755377593382 22.28660817344665 84.71174955368042
3.2136546546229927 -9.263162016899226 20.002318024635315
1.0699752667788094 -2.603866517556659 33.31665515899658
4.7496493192161315 -13.759816215056405 9.891473650932312
1.0204451415554694 -2.302884435650706 33.91196250915527
-9.543197093698641 27.579762022298645 95.41455745697021
-3.152390437546815 8.533435938355485 56.56946539878845
-6.972211698963628 19.499752653025283 79.14494276046753
-9.05184471581592 24.830375302932325 90.15502452850342
4.37743942426164 -12.384318614772202 13.527265191078186
5.4403943502954135 -15.45761907451561 8.019829988479614
-2.8655595340846656 8.461565986612753 56.21854305267334
-4.385969174990902 13.137465387174789 65.82960605621338
4.143933351224529 -11.719931529427582 15.11167287826538
3.97251556067615 -11.477948834650308 15.733973979949951
6.177795441914896 -17.73304217890827 2.9957453906536102
2.5499169587920774 -7.096689595677503 24.85772967338562
-9.130402062312143 26.249899116010937 92.76379108428955
-2.982628364485661 8.91535848876603 57.14838266372681
-3.547182977050129 9.888451725428762 59.32171583175659
-3.575685464368824 10.840678797888755 61.09130859375
6.247649163553585 -17.94350676043306 2.681722864508629
-6.186284895130248 17.919513500743104 75.69625854492188
4.473351473486235 -12.950998039149974 12.084179520606995
-2.508637271432117 6.818232381379776 52.99638390541077
5.782169247333714 -16.527788336257863 6.048831939697266
6.013702219104657 -17.18457536276333 4.183526337146759
-3.170607226140807 9.456728503947025 58.29724073410034
4.032525804087582 -11.661455671152707 15.355170965194702
2.8088999790378852 -7.667198855439364 23.295010328292847
-2.616825499479963 7.876500095347924 55.083500146865845
6.06325258867907 -17.29258530713545 3.9635534584522247
6.2841680590308275 -17.791095950051066 2.9174528270959854
-1.5699520189530882 5.210021426735137 49.538819789886475
6.2136682705395705 -17.696338994110505 3.022579327225685
-6.671554094645899 19.646873684887936 79.26994800567627
6.071350924005643 -17.37272072115279 3.8646014034748077
-7.616992919103632 21.757631692368825 83.58767509460449
-2.8784341740234805 7.869048077923313 55.07809281349182
5.826969177623034 -16.709075555194126 5.508389621973038
-7.586392608028636 21.038095803647952 82.32599973678589
6.1782901950234335 -17.802652480943312 2.871839851140976
3.9540950798112355 -11.379169430848007 15.93866229057312
-6.739772678051406 19.838653358064963 79.593186378479
-0.9634696299550488 3.0965671123780214 45.47806978225708
-9.99461099601671 28.180299278348294 96.78131103515625
5.671847048464663 -16.34461215197267 6.519233733415604
-5.200171523521748 14.446810511827547 68.849937915802
-12.217481291611819 34.17407259733368 108.99211406707764
5.7341300784460785 -16.330840826637054 6.502791345119476
6.05166537515429 -17.325282732214294 3.9089588820934296
-5.08370315607635 14.171130831120543 68.21295261383057
6.0520033733164675 -17.316680352946886 3.912636637687683
1.2452883099486618 -3.155195356233434 32.069681882858276
3.309494676916084 -9.52017509043921 19.408190846443176
5.852733992988925 -16.747641590000356 5.357412546873093
-0.636580370810977 2.3257445705696123 43.788546323776245
-7.18489291810015 20.998968445992503 81.91051483154297
6.103804574546865 -17.46360533693223 3.638395741581917
-8.728186380672337 24.226679729306408 88.86301517486572
5.801726064036828 -16.526974176302325 5.970014333724976
6.113574370454808 -17.40664478162494 3.829038441181183
2.724353425259188 -7.334888035481004 24.132996797561646
4.16550186073982 -11.7506387543712 15.081607103347778
2.057760692871829 -5.744151313375675 26.788628697395325
2.1864610938369684 -6.312189346122076 26.05537712574005
6.303248172884525 -18.142868739206044 2.3401933908462524
-7.915579901822024 23.073833785988047 86.25602960586548
0.8213668017762391 -1.8824827773611554 34.84135866165161
-1.063183553834116 3.042131579532693 45.43760061264038
};
\addplot+[only marks,scatter, 
mark=triangle*,
point meta=explicit,
point meta min={0}, point meta max={120},forget plot]
table[x=x,y=y,meta=w1]{
x y w1
-9.669794244731266 28.928154529293746 98.0391
-9.236794802650378 28.28424306249693 96.6953
-9.06643377184761 27.61991306530942 95.3543
-8.820519984191478 26.966772009178495 94.0172
-8.820978744246863 26.202148460629914 92.6847
-8.528213414693962 25.51924612850221 91.3574
-8.545108699824576 24.781735100140224 90.0357
-7.977270332073652 24.236452545290444 88.7198
-8.097459358879197 23.498026167080056 87.4074
-7.5210064215550805 22.973960833156116 86.0963
-7.585998186777977 22.227611911053643 84.7865
-7.073815552829354 21.755210735990502 83.4778
-7.233392344691892 20.96535069561643 82.1705
-6.97252784060527 20.319949831260455 80.8645
-6.491363420005653 19.76316821619155 79.5598
-6.264945059093594 19.15208208983562 78.2565
-6.362154839027946 18.430106839426685 76.9552
-6.111602831876736 17.792910016562033 75.6564
-5.589460807470645 17.29902245507544 74.3659
-5.508439545574762 16.58442356833325 73.0877
-5.343939096586253 15.953956410179407 71.823
-4.956761557746829 15.378885424365267 70.5721
-5.002314879657731 14.625580093169436 69.3346
-4.822713068580417 14.057821351748206 68.11
-4.544019993361941 13.476814347343018 66.8976
-4.123656917369345 12.966515329887375 65.6968
-4.1606765929593825 12.283348003767392 64.5071
-4.029853891319478 11.763089303120868 63.32809999999999
-3.5748264290189025 11.235526090218801 62.1595
-3.335433461073683 10.689792678193372 61.00119999999999
-3.2618113810522336 10.028170926142632 59.8528
-2.9646835239896125 9.561502163234744 58.7143
-8.716645488277333 26.549976233640034 57.5857
-8.930342266688712 25.892889051519873 56.4664
-8.956672130188434 25.831842154633552 55.3545
-8.513388214679352 25.568403404560854 54.2496
-8.392718167511777 25.58869400826777 53.1516
-8.561939172686532 25.418369503119646 52.0603
-8.357103556304283 25.435281451114435 50.9755
-8.399882339203609 25.4211762401741 49.897
-8.742777970017642 25.424461550785953 48.825
-8.568437639511469 25.06275474552363 47.7591
-8.38795281403216 24.468523274791032 46.6995
-8.178751710438299 24.098099006614007 45.646
-8.22197743031903 23.803375315068255 44.5985
-7.613534013959371 23.135581893848208 43.5572
-7.49728145360308 22.614037413849978 42.5218
-7.2321226151819875 22.073974119628577 41.4924
-7.448229619424343 21.55387331598852 40.4684
-6.916137587876041 21.381346382346642 39.446
-7.147761865492683 20.91882254151268 38.4245
-6.8896862409856 20.304953542050402 37.404
-6.81785740840643 20.146090580873647 36.3844
-6.693409108437473 19.44310277862441 35.3658
-6.472207120152785 19.250680233497857 34.3481
-6.585380888169691 19.159405502770696 33.3314
-6.214607656145599 19.078843271494108 32.3155
-6.299935563733482 18.92105842572491 31.3006
-6.208702152871078 19.299485361295126 30.2867
-6.23559968231411 18.35208390718934 29.2822
-6.081195567366661 18.245599779948197 28.2908
-5.946128751722779 18.063131071195162 27.3111
-5.989581497826853 17.226723366343546 26.3423
-5.679497961391237 17.577726285996828 25.3838
-5.770783016577594 17.606335461575018 24.4351
-5.899817476673117 17.532917536420335 23.4959
-5.663098909820781 17.277372534691533 22.5658
-5.698079840945574 16.810009526568525 21.6446
-5.412011146361799 16.217206516243646 20.7322
-5.190210362696726 15.978502462858861 19.8284
-5.473386273533323 15.61547287097987 18.933
-5.2335364966275355 15.298164581386608 18.046
-5.0338143879138215 14.563201461707397 17.1672
-4.7414476807236845 14.310697290950815 16.2965
-4.533047352025371 13.65239156310649 15.433800000000002
-4.3626871488933645 13.490358209135536 14.5791
-4.248536870778008 13.226470591290234 13.7324
-4.375946939102561 12.944108966627404 12.8934
-4.379979300936216 12.788776912593601 12.0622
-4.0500933948560345 12.90526230009633 11.2386
-4.218442911137544 12.48774971809091 10.4227
-4.3868901764450925 12.769432446180147 9.6143
-4.022178797514061 12.584145896442022 8.8134
-4.309725218351082 12.549145436381702 8.0199
-3.9466810666309153 12.171813198114481 7.2337
-3.7809217632905816 11.963098659683808 6.4549
-4.082469092626036 11.735859727616504 5.6833
-3.9598680459212603 11.399051020458845 4.9188
-3.824957150319627 11.279050057054224 4.1615
-3.8080465050569603 10.996328607647218 3.4089
-3.559727726961544 11.224115985696992 2.6566
-3.822841332516933 10.913384363610534 1.905
-3.8629426050686453 11.180080811156088 1.1541
-3.850274859063502 11.431848029916635 0.4038
-4.048916553367366 12.351260106690907 -0.3457
};

\addplot+[only marks,scatter, 
mark=o, draw=black]
table[x=x,y=y]{
	x y
	100 100
};\addlegendentry{In-distribution}

\addplot+[only marks,scatter, 
mark=triangle, draw=black]
table[x=x,y=y]{
x y
100 100
};\addlegendentry{Adversarial}
%\legend{In-distribution, Adversarial}

\end{axis}

\end{tikzpicture}%

%% file: figures/attackPerception.tex
% This file was created by matlab2tikz.
%
%The latest updates can be retrieved from
%  http://www.mathworks.com/matlabcentral/fileexchange/22022-matlab2tikz-matlab2tikz
%where you can also make suggestions and rate matlab2tikz.
%
\begin{tikzpicture}
\pgfplotsset{
	scale only axis,
	width=7.0cm,
	height=2.0cm,
	%major grid style={dotted, gray}
}
\begin{groupplot}[
group style={
	group size = 1 by 5,
	%x descriptions at = edge bottom,
	vertical sep=2pt,
},		
label style={font=\footnotesize},
tick label style={font=\footnotesize},
legend style={font=\scriptsize},
y label style={at = {(0,0.5)}, yshift=1.5em, align=center},
]
\nextgroupplot[
xmin = 0, xmax=5.0,
ymin = 0, ymax=100,
xtick = {0,1,2,3,4,5,6,7},
xticklabels = {0, 20, 40, 60, 80, 100, 120, 140},
ytick = {0, 20, 40, 60, 80, 100},
xticklabel pos=top,
xlabel={Simulation step},
ylabel={Distance($\SI{}{\meter}$)},
legend cell align={left},
]
\addplot [color=red]
table[row sep=crcr]{%
0.0 99.3667 \\
0.05 98.0073 \\
0.1 96.6453 \\
0.15 95.2825 \\
0.2 93.9213 \\
0.25 92.5631 \\
0.3 91.209 \\
0.35 89.8597 \\
0.4 88.5158 \\
0.45 87.1751 \\
0.5 85.8358 \\
0.55 84.4978 \\
0.6 83.1612 \\
0.65 81.826 \\
0.7 80.4926 \\
0.75 79.1618 \\
0.8 77.8349 \\
0.85 76.5145 \\
0.9 75.2037 \\
0.95 73.9069 \\
1.0 72.6267 \\
1.05 71.3635 \\
1.1 70.1169 \\
1.15 68.8857 \\
1.2 67.6684 \\
1.25 66.4637 \\
1.3 65.2705 \\
1.35 64.0876 \\
1.4 62.9131 \\
1.45 61.7464 \\
1.5 60.5873 \\
1.55 59.4354 \\
1.6 58.2906 \\
1.65 57.1526 \\
1.7 56.0214 \\
1.75 54.8967 \\
1.8 53.7787 \\
1.85 52.6671 \\
1.9 51.562 \\
1.95 50.4632 \\
2.0 49.3707 \\
2.05 48.2846 \\
2.1 47.2047 \\
2.15 46.131 \\
2.2 45.0634 \\
2.25 44.002 \\
2.3 42.9467 \\
2.35 41.8975 \\
2.4 40.8542 \\
2.45 39.8166 \\
2.5 38.7806 \\
2.55 37.7455 \\
2.6 36.7114 \\
2.65 35.6784 \\
2.7 34.6466 \\
2.75 33.6162 \\
2.8 32.5878 \\
2.85 31.562 \\
2.9 30.5397 \\
2.95 29.522 \\
3.0 28.5168 \\
3.05 27.5282 \\
3.1 26.5553 \\
3.15 25.5966 \\
3.2 24.6507 \\
3.25 23.7165 \\
3.3 22.7933 \\
3.35 21.8807 \\
3.4 20.978 \\
3.45 20.0849 \\
3.5 19.201 \\
3.55 18.3262 \\
3.6 17.4603 \\
3.65 16.6033 \\
3.7 15.7552 \\
3.75 14.916 \\
3.8 14.086 \\
3.85 13.2653 \\
3.9 12.4543 \\
3.95 11.653 \\
4.0 10.8619 \\
4.05 10.0814 \\
4.1 9.3119 \\
4.15 8.5543 \\
4.2 7.8095 \\
4.25 7.0779 \\
4.3 6.3602 \\
4.35 5.6572 \\
4.4 4.9699 \\
4.45 4.2987 \\
4.5 3.642 \\
4.55 2.9982 \\
4.6 2.3676 \\
4.65 1.7496 \\
4.7 1.145 \\
4.75 0.5542 \\
4.8 -0.0241 \\
};\addlegendentry{Ground truth distance}

\addplot [color=black, dashed]
table[row sep=crcr]{%
0.0 99.5843 \\
0.05 98.1544 \\
0.1 96.7037 \\
0.15 95.2094 \\
0.2 93.9909 \\
0.25 92.6161 \\
0.3 91.2433 \\
0.35 89.7678 \\
0.4 88.3604 \\
0.45 87.2311 \\
0.5 85.9331 \\
0.55 84.5674 \\
0.6 83.2584 \\
0.65 81.9354 \\
0.7 80.5693 \\
0.75 79.241 \\
0.8 78.0017 \\
0.85 76.6508 \\
0.9 75.3434 \\
0.95 73.8842 \\
1.0 72.615 \\
1.05 71.3916 \\
1.1 70.0719 \\
1.15 68.9086 \\
1.2 67.6751 \\
1.25 79.4519 \\
1.3 78.1866 \\
1.35 76.8476 \\
1.4 75.7235 \\
1.45 74.5605 \\
1.5 73.6244 \\
1.55 72.6528 \\
1.6 71.4986 \\
1.65 70.2083 \\
1.7 68.9156 \\
1.75 67.9422 \\
1.8 67.0869 \\
1.85 66.0914 \\
1.9 65.1259 \\
1.95 64.3669 \\
2.0 63.2236 \\
2.05 62.3143 \\
2.1 60.8136 \\
2.15 59.7756 \\
2.2 58.7536 \\
2.25 57.7267 \\
2.3 56.7323 \\
2.35 55.901 \\
2.4 54.8448 \\
2.45 53.7694 \\
2.5 52.7493 \\
2.55 51.4159 \\
2.6 50.4587 \\
2.65 49.3975 \\
2.7 48.3439 \\
2.75 47.7325 \\
2.8 46.9151 \\
2.85 46.0039 \\
2.9 44.9973 \\
2.95 43.94 \\
3.0 43.0453 \\
3.05 42.0241 \\
3.1 40.9364 \\
3.15 39.8878 \\
3.2 39.0718 \\
3.25 38.1735 \\
3.3 37.9257 \\
3.35 37.0186 \\
3.4 36.1035 \\
3.45 35.0207 \\
3.5 34.0216 \\
3.55 32.9357 \\
3.6 32.059 \\
3.65 30.9534 \\
3.7 29.9249 \\
3.75 28.8789 \\
3.8 28.3016 \\
3.85 27.5362 \\
3.9 26.6272 \\
3.95 25.9289 \\
4.0 25.2803 \\
4.05 24.2856 \\
4.1 23.8952 \\
4.15 23.0834 \\
4.2 22.5387 \\
4.25 21.6773 \\
4.3 20.9852 \\
4.35 20.6356 \\
4.4 19.9955 \\
4.45 19.2486 \\
4.5 18.9299 \\
4.55 18.2713 \\
4.6 17.7601 \\
4.65 17.1461 \\
4.7 17.8415 \\
4.75 18.7486 \\
4.8 19.612 \\
};\addplot [red,dashed,line width=1pt]  coordinates { (1.20,0)(1.20,100) };
\addlegendentry{Predicted distance}

\nextgroupplot[
xmin = 0, xmax=5.0,
ymin = 0, ymax=25,
ytick = {0, 10, 20},
ylabel={Prediction error($\SI{}{\meter}$)},
xticklabels={,,},
]
\addplot [color=blue]
table[row sep=crcr]{%
0.0 0.21760000000000446 \\
0.05 0.14709999999999468 \\
0.1 0.05839999999999179 \\
0.15 0.07309999999999661 \\
0.2 0.06959999999999411 \\
0.25 0.05299999999999727 \\
0.3 0.034300000000001774 \\
0.35 0.09190000000000964 \\
0.4 0.1554000000000002 \\
0.45 0.055999999999997385 \\
0.5 0.09729999999998995 \\
0.55 0.06960000000000832 \\
0.6 0.09720000000000084 \\
0.65 0.10940000000000794 \\
0.7 0.07670000000000243 \\
0.75 0.07920000000000016 \\
0.8 0.16679999999999495 \\
0.85 0.13630000000000564 \\
0.9 0.13970000000000482 \\
0.95 0.022699999999986176 \\
1.0 0.011700000000004707 \\
1.05 0.028099999999994907 \\
1.1 0.045000000000001705 \\
1.15 0.022900000000007026 \\
1.2 0.006699999999995043 \\
1.25 12.988199999999992 \\
1.3 12.9161 \\
1.35 12.760000000000005 \\
1.4 12.810400000000001 \\
1.45 12.814100000000003 \\
1.5 13.037099999999995 \\
1.55 13.217399999999998 \\
1.6 13.207999999999998 \\
1.65 13.055699999999995 \\
1.7 12.894199999999998 \\
1.75 13.045499999999997 \\
1.8 13.3082 \\
1.85 13.424299999999995 \\
1.9 13.563900000000004 \\
1.95 13.9037 \\
2.0 13.852899999999998 \\
2.05 14.029700000000005 \\
2.1 13.608899999999998 \\
2.15 13.644599999999997 \\
2.2 13.690199999999997 \\
2.25 13.724699999999999 \\
2.3 13.785600000000002 \\
2.35 14.003500000000003 \\
2.4 13.9906 \\
2.45 13.952799999999996 \\
2.5 13.968699999999998 \\
2.55 13.6704 \\
2.6 13.747300000000003 \\
2.65 13.719099999999997 \\
2.7 13.697299999999998 \\
2.75 14.116300000000003 \\
2.8 14.327300000000001 \\
2.85 14.4419 \\
2.9 14.457600000000003 \\
2.95 14.418 \\
3.0 14.528499999999998 \\
3.05 14.495899999999999 \\
3.1 14.3811 \\
3.15 14.2912 \\
3.2 14.421100000000003 \\
3.25 14.456999999999997 \\
3.3 15.1324 \\
3.35 15.137899999999998 \\
3.4 15.125499999999995 \\
3.45 14.935799999999997 \\
3.5 14.820599999999999 \\
3.55 14.609499999999997 \\
3.6 14.598699999999997 \\
3.65 14.350099999999998 \\
3.7 14.1697 \\
3.75 13.962900000000001 \\
3.8 14.2156 \\
3.85 14.270900000000001 \\
3.9 14.172899999999998 \\
3.95 14.275899999999998 \\
4.0 14.4184 \\
4.05 14.204199999999998 \\
4.1 14.5833 \\
4.15 14.529100000000001 \\
4.2 14.729199999999999 \\
4.25 14.5994 \\
4.3 14.625 \\
4.35 14.9784 \\
4.4 15.0256 \\
4.45 14.9499 \\
4.5 15.2879 \\
4.55 15.2731 \\
4.6 15.392500000000002 \\
4.65 15.3965 \\
4.7 16.6965 \\
4.75 18.194399999999998 \\
4.8 19.6361 \\
};
\addplot [red,dashed,line width=1pt]  coordinates { (1.20,0)(1.20,70) };

% start the second plot
\nextgroupplot[
xmin = 0, xmax=5.0,
ymin = 0, ymax=35,
ytick = {0, 10, 20, 30},
xticklabels={,,},
ylabel={Velocity($\SI[per-mode=symbol]{}{\meter\per\second}$)},
]
\addplot [color=blue]
table[row sep=crcr]{%
0.0 27.1409 \\
0.05 27.2098 \\
0.1 27.2545 \\
0.15 27.2545 \\
0.2 27.2143 \\
0.25 27.1465 \\
0.3 27.0594 \\
0.35 26.9589 \\
0.4 26.849 \\
0.45 26.805 \\
0.5 26.7777 \\
0.55 26.7509 \\
0.6 26.7232 \\
0.65 26.6922 \\
0.7 26.6533 \\
0.75 26.5961 \\
0.8 26.5047 \\
0.85 26.3596 \\
0.9 26.1498 \\
0.95 25.8417 \\
1.0 25.5056 \\
1.05 25.1644 \\
1.1 24.8361 \\
1.15 24.5345 \\
1.2 24.2618 \\
1.25 24.0153 \\
1.3 23.7898 \\
1.35 23.5972 \\
1.4 23.4324 \\
1.45 23.2773 \\
1.5 23.1292 \\
1.55 22.9863 \\
1.6 22.8473 \\
1.65 22.7114 \\
1.7 22.5777 \\
1.75 22.4459 \\
1.8 22.3156 \\
1.85 22.1866 \\
1.9 22.0587 \\
1.95 21.9318 \\
2.0 21.8057 \\
2.05 21.6805 \\
2.1 21.556 \\
2.15 21.4323 \\
2.2 21.3092 \\
2.25 21.1869 \\
2.3 21.0652 \\
2.35 20.9442 \\
2.4 20.8237 \\
2.45 20.7297 \\
2.5 20.7087 \\
2.55 20.6895 \\
2.6 20.6699 \\
2.65 20.6486 \\
2.7 20.6243 \\
2.75 20.5937 \\
2.8 20.5507 \\
2.85 20.4943 \\
2.9 20.4204 \\
2.95 20.3203 \\
3.0 20.0058 \\
3.05 19.6791 \\
3.1 19.3705 \\
3.15 19.0941 \\
3.2 18.8435 \\
3.25 18.6111 \\
3.3 18.3925 \\
3.35 18.185 \\
3.4 17.9883 \\
3.45 17.7996 \\
3.5 17.616 \\
3.55 17.4356 \\
3.6 17.257 \\
3.65 17.079 \\
3.7 16.9013 \\
3.75 16.7224 \\
3.8 16.5394 \\
3.85 16.3492 \\
3.9 16.156 \\
3.95 15.9598 \\
4.0 15.7544 \\
4.05 15.5391 \\
4.1 15.3149 \\
4.15 15.0709 \\
4.2 14.8142 \\
4.25 14.5435 \\
4.3 14.2619 \\
4.35 13.9628 \\
4.4 13.643 \\
4.45 13.3245 \\
4.5 13.0512 \\
4.55 12.7845 \\
4.6 12.5305 \\
4.65 12.2702 \\
4.7 12.0013 \\
4.75 11.7208 \\
4.8 11.5111 \\
};\addplot [red,dashed,line width=1pt]  coordinates { (1.20,0)(1.20,35) };

\nextgroupplot[
xmin = 0, xmax=5.0,
ymin = 0.0, ymax=1.1,
ytick = {0, 0.5, 1.0},
xticklabels={,,},
ylabel={$p$ value},
]
\addplot [color=blue, only marks, mark size=0.5pt]
table[row sep=crcr]{%
0.0 0.6826137689614936 \\
0.05 0.6592765460910153 \\
0.1 0.5414235705950992 \\
0.15 0.8284714119019837 \\
0.2 0.9317386231038506 \\
0.25 0.9679113185530922 \\
0.3 0.9474912485414235 \\
0.35 0.8430571761960327 \\
0.4 0.9031505250875146 \\
0.45 0.9579929988331389 \\
0.5 0.9346557759626605 \\
0.55 0.7712952158693116 \\
0.6 0.7450408401400233 \\
0.65 0.5997666277712952 \\
0.7 0.7357059509918319 \\
0.75 0.9154025670945157 \\
0.8 0.9235705950991832 \\
0.85 0.8879813302217037 \\
0.9 0.9107351225204201 \\
0.95 0.9358226371061843 \\
1.0 0.8028004667444574 \\
1.05 0.8885647607934657 \\
1.1 0.8652275379229871 \\
1.15 0.808051341890315 \\
1.2 0.8547257876312719 \\
1.25 0.0 \\
1.3 0.0 \\
1.35 0.0 \\
1.4 0.0 \\
1.45 0.0 \\
1.5 0.0 \\
1.55 0.0 \\
1.6 0.0 \\
1.65 0.0 \\
1.7 0.0 \\
1.75 0.0 \\
1.8 0.0 \\
1.85 0.0 \\
1.9 0.0 \\
1.95 0.0 \\
2.0 0.0 \\
2.05 0.0 \\
2.1 0.0 \\
2.15 0.0 \\
2.2 0.0 \\
2.25 0.0 \\
2.3 0.0 \\
2.35 0.0 \\
2.4 0.0 \\
2.45 0.0 \\
2.5 0.0 \\
2.55 0.0 \\
2.6 0.0 \\
2.65 0.0 \\
2.7 0.0 \\
2.75 0.0 \\
2.8 0.0 \\
2.85 0.0 \\
2.9 0.0 \\
2.95 0.0 \\
3.0 0.0 \\
3.05 0.0 \\
3.1 0.0 \\
3.15 0.0 \\
3.2 0.0 \\
3.25 0.0 \\
3.3 0.0 \\
3.35 0.0 \\
3.4 0.0 \\
3.45 0.0 \\
3.5 0.0 \\
3.55 0.0 \\
3.6 0.0 \\
3.65 0.0 \\
3.7 0.0 \\
3.75 0.0 \\
3.8 0.0 \\
3.85 0.0 \\
3.9 0.0 \\
3.95 0.0 \\
4.0 0.0 \\
4.05 0.0 \\
4.1 0.0 \\
4.15 0.0 \\
4.2 0.0 \\
4.25 0.0 \\
4.3 0.0 \\
4.35 0.0 \\
4.4 0.0 \\
4.45 0.0 \\
4.5 0.0 \\
4.55 0.0 \\
4.6 0.0 \\
4.65 0.0 \\
4.7 0.0 \\
4.75 0.0 \\
4.8 0.0 \\
0.0 0.6516919486581096 \\
0.05 0.6277712952158693 \\
0.1 0.5268378063010501 \\
0.15 0.8284714119019837 \\
0.2 0.9171528588098016 \\
0.25 0.8523920653442241 \\
0.3 0.8879813302217037 \\
0.35 0.8640606767794633 \\
0.4 0.8506417736289381 \\
0.45 0.9434072345390898 \\
0.5 0.941656942823804 \\
0.55 0.883313885647608 \\
0.6 0.7082847141190198 \\
0.65 0.5950991831971995 \\
0.7 0.7456242707117853 \\
0.75 0.912485414235706 \\
0.8 0.9346557759626605 \\
0.85 0.822053675612602 \\
0.9 0.9194865810968494 \\
0.95 0.8984830805134189 \\
1.0 0.9194865810968494 \\
1.05 0.9031505250875146 \\
1.1 0.868728121353559 \\
1.15 0.7473745624270711 \\
1.2 0.8453908984830805 \\
1.25 0.0 \\
1.3 0.0 \\
1.35 0.0 \\
1.4 0.0 \\
1.45 0.0 \\
1.5 0.0 \\
1.55 0.0 \\
1.6 0.0 \\
1.65 0.0 \\
1.7 0.0 \\
1.75 0.0 \\
1.8 0.0 \\
1.85 0.0 \\
1.9 0.0 \\
1.95 0.0 \\
2.0 0.0 \\
2.05 0.0 \\
2.1 0.0 \\
2.15 0.0 \\
2.2 0.0 \\
2.25 0.0 \\
2.3 0.0 \\
2.35 0.0 \\
2.4 0.0 \\
2.45 0.0 \\
2.5 0.0 \\
2.55 0.0 \\
2.6 0.0 \\
2.65 0.0 \\
2.7 0.0 \\
2.75 0.0 \\
2.8 0.0 \\
2.85 0.0 \\
2.9 0.0 \\
2.95 0.0 \\
3.0 0.0 \\
3.05 0.0 \\
3.1 0.0 \\
3.15 0.0 \\
3.2 0.0 \\
3.25 0.0 \\
3.3 0.0 \\
3.35 0.0 \\
3.4 0.0 \\
3.45 0.0 \\
3.5 0.0 \\
3.55 0.0 \\
3.6 0.0 \\
3.65 0.0 \\
3.7 0.0 \\
3.75 0.0 \\
3.8 0.0 \\
3.85 0.0 \\
3.9 0.0 \\
3.95 0.0 \\
4.0 0.0 \\
4.05 0.0 \\
4.1 0.0 \\
4.15 0.0 \\
4.2 0.0 \\
4.25 0.0 \\
4.3 0.0 \\
4.35 0.0 \\
4.4 0.0 \\
4.45 0.0 \\
4.5 0.0 \\
4.55 0.0 \\
4.6 0.0 \\
4.65 0.0 \\
4.7 0.0 \\
4.75 0.0 \\
4.8 0.0 \\
0.0 0.6487747957993 \\
0.05 0.5670945157526254 \\
0.1 0.7222870478413068 \\
0.15 0.7922987164527422 \\
0.2 0.9404900816802801 \\
0.25 0.9486581096849475 \\
0.3 0.9550758459743292 \\
0.35 0.8868144690781797 \\
0.4 0.9084014002333722 \\
0.45 0.9667444574095683 \\
0.5 0.9171528588098016 \\
0.55 0.9031505250875146 \\
0.6 0.7444574095682615 \\
0.65 0.5455075845974329 \\
0.7 0.707117852975496 \\
0.75 0.8704784130688448 \\
0.8 0.9422403733955659 \\
0.85 0.868144690781797 \\
0.9 0.8121353558926487 \\
0.95 0.9358226371061843 \\
1.0 0.8879813302217037 \\
1.05 0.8809801633605601 \\
1.1 0.8442240373395566 \\
1.15 0.8004667444574096 \\
1.2 0.8547257876312719 \\
1.25 0.0 \\
1.3 0.0 \\
1.35 0.0 \\
1.4 0.0 \\
1.45 0.0 \\
1.5 0.0 \\
1.55 0.0 \\
1.6 0.0 \\
1.65 0.0 \\
1.7 0.0 \\
1.75 0.0 \\
1.8 0.0 \\
1.85 0.0 \\
1.9 0.0 \\
1.95 0.0 \\
2.0 0.0 \\
2.05 0.0 \\
2.1 0.0 \\
2.15 0.0 \\
2.2 0.0 \\
2.25 0.0 \\
2.3 0.0 \\
2.35 0.0 \\
2.4 0.0 \\
2.45 0.0 \\
2.5 0.0 \\
2.55 0.0 \\
2.6 0.0 \\
2.65 0.0 \\
2.7 0.0 \\
2.75 0.0 \\
2.8 0.0 \\
2.85 0.0 \\
2.9 0.0 \\
2.95 0.0 \\
3.0 0.0 \\
3.05 0.0 \\
3.1 0.0 \\
3.15 0.0 \\
3.2 0.0 \\
3.25 0.0 \\
3.3 0.0 \\
3.35 0.0 \\
3.4 0.0 \\
3.45 0.0 \\
3.5 0.0 \\
3.55 0.0 \\
3.6 0.0 \\
3.65 0.0 \\
3.7 0.0 \\
3.75 0.0 \\
3.8 0.0 \\
3.85 0.0 \\
3.9 0.0 \\
3.95 0.0 \\
4.0 0.0 \\
4.05 0.0 \\
4.1 0.0 \\
4.15 0.0 \\
4.2 0.0 \\
4.25 0.0 \\
4.3 0.0 \\
4.35 0.0 \\
4.4 0.0 \\
4.45 0.0 \\
4.5 0.0 \\
4.55 0.0 \\
4.6 0.0 \\
4.65 0.0 \\
4.7 0.0 \\
4.75 0.0 \\
4.8 0.0 \\
0.0 0.5635939323220537 \\
0.05 0.6219369894982497 \\
0.1 0.6721120186697783 \\
0.15 0.8319719953325554 \\
0.2 0.9375729288214703 \\
0.25 0.9673278879813302 \\
0.3 0.9043173862310385 \\
0.35 0.8623103850641773 \\
0.4 0.9597432905484247 \\
0.45 0.9649941656942823 \\
0.5 0.9084014002333722 \\
0.55 0.8768961493582264 \\
0.6 0.7380396732788799 \\
0.65 0.5921820303383898 \\
0.7 0.5612602100350058 \\
0.75 0.9119019836639439 \\
0.8 0.9101516919486581 \\
0.85 0.883313885647608 \\
0.9 0.9060676779463245 \\
0.95 0.9428238039673279 \\
1.0 0.8739789964994165 \\
1.05 0.897899649941657 \\
1.1 0.823803967327888 \\
1.15 0.8168028004667446 \\
1.2 0.8378063010501751 \\
1.25 0.0 \\
1.3 0.0 \\
1.35 0.0 \\
1.4 0.0 \\
1.45 0.0 \\
1.5 0.0 \\
1.55 0.0 \\
1.6 0.0 \\
1.65 0.0 \\
1.7 0.0 \\
1.75 0.0 \\
1.8 0.0 \\
1.85 0.0 \\
1.9 0.0 \\
1.95 0.0 \\
2.0 0.0 \\
2.05 0.0 \\
2.1 0.0 \\
2.15 0.0 \\
2.2 0.0 \\
2.25 0.0 \\
2.3 0.0 \\
2.35 0.0 \\
2.4 0.0 \\
2.45 0.0 \\
2.5 0.0 \\
2.55 0.0 \\
2.6 0.0 \\
2.65 0.0 \\
2.7 0.0 \\
2.75 0.0 \\
2.8 0.0 \\
2.85 0.0 \\
2.9 0.0 \\
2.95 0.0 \\
3.0 0.0 \\
3.05 0.0 \\
3.1 0.0 \\
3.15 0.0 \\
3.2 0.0 \\
3.25 0.0 \\
3.3 0.0 \\
3.35 0.0 \\
3.4 0.0 \\
3.45 0.0 \\
3.5 0.0 \\
3.55 0.0 \\
3.6 0.0 \\
3.65 0.0 \\
3.7 0.0 \\
3.75 0.0 \\
3.8 0.0 \\
3.85 0.0 \\
3.9 0.0 \\
3.95 0.0 \\
4.0 0.0 \\
4.05 0.0 \\
4.1 0.0 \\
4.15 0.0 \\
4.2 0.0 \\
4.25 0.0 \\
4.3 0.0 \\
4.35 0.0 \\
4.4 0.0 \\
4.45 0.0 \\
4.5 0.0 \\
4.55 0.0 \\
4.6 0.0 \\
4.65 0.0 \\
4.7 0.0 \\
4.75 0.0 \\
4.8 0.0 \\
0.0 0.6826137689614936 \\
0.05 0.4813302217036173 \\
0.1 0.5863477246207701 \\
0.15 0.8016336056009336 \\
0.2 0.9008168028004667 \\
0.25 0.9644107351225205 \\
0.3 0.9101516919486581 \\
0.35 0.9113185530921819 \\
0.4 0.9060676779463245 \\
0.45 0.9253208868144691 \\
0.5 0.8955659276546091 \\
0.55 0.9066511085180864 \\
0.6 0.7438739789964994 \\
0.65 0.5997666277712952 \\
0.7 0.7497082847141191 \\
0.75 0.8494749124854143 \\
0.8 0.9329054842473745 \\
0.85 0.8424737456242707 \\
0.9 0.9171528588098016 \\
0.95 0.9107351225204201 \\
1.0 0.9194865810968494 \\
1.05 0.8319719953325554 \\
1.1 0.8459743290548425 \\
1.15 0.8016336056009336 \\
1.2 0.8693115519253208 \\
1.25 0.0 \\
1.3 0.0 \\
1.35 0.0 \\
1.4 0.0 \\
1.45 0.0 \\
1.5 0.0 \\
1.55 0.0 \\
1.6 0.0 \\
1.65 0.0 \\
1.7 0.0 \\
1.75 0.0 \\
1.8 0.0 \\
1.85 0.0 \\
1.9 0.0 \\
1.95 0.0 \\
2.0 0.0 \\
2.05 0.0 \\
2.1 0.0 \\
2.15 0.0 \\
2.2 0.0 \\
2.25 0.0 \\
2.3 0.0 \\
2.35 0.0 \\
2.4 0.0 \\
2.45 0.0 \\
2.5 0.0 \\
2.55 0.0 \\
2.6 0.0 \\
2.65 0.0 \\
2.7 0.0 \\
2.75 0.0 \\
2.8 0.0 \\
2.85 0.0 \\
2.9 0.0 \\
2.95 0.0 \\
3.0 0.0 \\
3.05 0.0 \\
3.1 0.0 \\
3.15 0.0 \\
3.2 0.0 \\
3.25 0.0 \\
3.3 0.0 \\
3.35 0.0 \\
3.4 0.0 \\
3.45 0.0 \\
3.5 0.0 \\
3.55 0.0 \\
3.6 0.0 \\
3.65 0.0 \\
3.7 0.0 \\
3.75 0.0 \\
3.8 0.0 \\
3.85 0.0 \\
3.9 0.0 \\
3.95 0.0 \\
4.0 0.0 \\
4.05 0.0 \\
4.1 0.0 \\
4.15 0.0 \\
4.2 0.0 \\
4.25 0.0 \\
4.3 0.0 \\
4.35 0.0 \\
4.4 0.0 \\
4.45 0.0 \\
4.5 0.0 \\
4.55 0.0 \\
4.6 0.0 \\
4.65 0.0 \\
4.7 0.0 \\
4.75 0.0 \\
4.8 0.0 \\
0.0 0.6067677946324388 \\
0.05 0.5455075845974329 \\
0.1 0.6277712952158693 \\
0.15 0.7520420070011669 \\
0.2 0.9194865810968494 \\
0.25 0.9579929988331389 \\
0.3 0.9130688448074679 \\
0.35 0.8704784130688448 \\
0.4 0.956242707117853 \\
0.45 0.9597432905484247 \\
0.5 0.9568261376896149 \\
0.55 0.9072345390898484 \\
0.6 0.6837806301050174 \\
0.65 0.5198366394399067 \\
0.7 0.7444574095682615 \\
0.75 0.8943990665110851 \\
0.8 0.7800466744457409 \\
0.85 0.8885647607934657 \\
0.9 0.8868144690781797 \\
0.95 0.9259043173862311 \\
1.0 0.8885647607934657 \\
1.05 0.9014002333722287 \\
1.1 0.8418903150525088 \\
1.15 0.8162193698949826 \\
1.2 0.8547257876312719 \\
1.25 0.0 \\
1.3 0.0 \\
1.35 0.0 \\
1.4 0.0 \\
1.45 0.0 \\
1.5 0.0 \\
1.55 0.0 \\
1.6 0.0 \\
1.65 0.0 \\
1.7 0.0 \\
1.75 0.0 \\
1.8 0.0 \\
1.85 0.0 \\
1.9 0.0 \\
1.95 0.0 \\
2.0 0.0 \\
2.05 0.0 \\
2.1 0.0 \\
2.15 0.0 \\
2.2 0.0 \\
2.25 0.0 \\
2.3 0.0 \\
2.35 0.0 \\
2.4 0.0 \\
2.45 0.0 \\
2.5 0.0 \\
2.55 0.0 \\
2.6 0.0 \\
2.65 0.0 \\
2.7 0.0 \\
2.75 0.0 \\
2.8 0.0 \\
2.85 0.0 \\
2.9 0.0 \\
2.95 0.0 \\
3.0 0.0 \\
3.05 0.0 \\
3.1 0.0 \\
3.15 0.0 \\
3.2 0.0 \\
3.25 0.0 \\
3.3 0.0 \\
3.35 0.0 \\
3.4 0.0 \\
3.45 0.0 \\
3.5 0.0 \\
3.55 0.0 \\
3.6 0.0 \\
3.65 0.0 \\
3.7 0.0 \\
3.75 0.0 \\
3.8 0.0 \\
3.85 0.0 \\
3.9 0.0 \\
3.95 0.0 \\
4.0 0.0 \\
4.05 0.0 \\
4.1 0.0 \\
4.15 0.0 \\
4.2 0.0 \\
4.25 0.0 \\
4.3 0.0 \\
4.35 0.0 \\
4.4 0.0 \\
4.45 0.0 \\
4.5 0.0 \\
4.55 0.0 \\
4.6 0.0 \\
4.65 0.0 \\
4.7 0.0 \\
4.75 0.0 \\
4.8 0.0 \\
0.0 0.690781796966161 \\
0.05 0.5921820303383898 \\
0.1 0.7386231038506418 \\
0.15 0.8150525087514586 \\
0.2 0.7444574095682615 \\
0.25 0.9585764294049008 \\
0.3 0.8716452742123687 \\
0.35 0.8646441073512253 \\
0.4 0.9317386231038506 \\
0.45 0.9597432905484247 \\
0.5 0.7642940490081681 \\
0.55 0.9002333722287048 \\
0.6 0.7438739789964994 \\
0.65 0.4428238039673279 \\
0.7 0.7438739789964994 \\
0.75 0.8599766627771295 \\
0.8 0.9410735122520419 \\
0.85 0.8716452742123687 \\
0.9 0.9031505250875146 \\
0.95 0.9212368728121354 \\
1.0 0.8996499416569428 \\
1.05 0.8844807467911319 \\
1.1 0.8623103850641773 \\
1.15 0.8121353558926487 \\
1.2 0.8634772462077013 \\
1.25 0.0 \\
1.3 0.0 \\
1.35 0.0 \\
1.4 0.0 \\
1.45 0.0 \\
1.5 0.0 \\
1.55 0.0 \\
1.6 0.0 \\
1.65 0.0 \\
1.7 0.0 \\
1.75 0.0 \\
1.8 0.0 \\
1.85 0.0 \\
1.9 0.0 \\
1.95 0.0 \\
2.0 0.0 \\
2.05 0.0 \\
2.1 0.0 \\
2.15 0.0 \\
2.2 0.0 \\
2.25 0.0 \\
2.3 0.0 \\
2.35 0.0 \\
2.4 0.0 \\
2.45 0.0 \\
2.5 0.0 \\
2.55 0.0 \\
2.6 0.0 \\
2.65 0.0 \\
2.7 0.0 \\
2.75 0.0 \\
2.8 0.0 \\
2.85 0.0 \\
2.9 0.0 \\
2.95 0.0 \\
3.0 0.0 \\
3.05 0.0 \\
3.1 0.0 \\
3.15 0.0 \\
3.2 0.0 \\
3.25 0.0 \\
3.3 0.0 \\
3.35 0.0 \\
3.4 0.0 \\
3.45 0.0 \\
3.5 0.0 \\
3.55 0.0 \\
3.6 0.0 \\
3.65 0.0 \\
3.7 0.0 \\
3.75 0.0 \\
3.8 0.0 \\
3.85 0.0 \\
3.9 0.0 \\
3.95 0.0 \\
4.0 0.0 \\
4.05 0.0 \\
4.1 0.0 \\
4.15 0.0 \\
4.2 0.0 \\
4.25 0.0 \\
4.3 0.0 \\
4.35 0.0 \\
4.4 0.0 \\
4.45 0.0 \\
4.5 0.0 \\
4.55 0.0 \\
4.6 0.0 \\
4.65 0.0 \\
4.7 0.0 \\
4.75 0.0 \\
4.8 0.0 \\
0.0 0.6563593932322053 \\
0.05 0.5851808634772462 \\
0.1 0.5542590431738623 \\
0.15 0.7479579929988331 \\
0.2 0.9317386231038506 \\
0.25 0.9673278879813302 \\
0.3 0.8786464410735122 \\
0.35 0.9072345390898484 \\
0.4 0.9095682613768962 \\
0.45 0.9644107351225205 \\
0.5 0.9218203033838974 \\
0.55 0.8990665110851809 \\
0.6 0.677946324387398 \\
0.65 0.5478413068844807 \\
0.7 0.7502917152858809 \\
0.75 0.8494749124854143 \\
0.8 0.912485414235706 \\
0.85 0.8885647607934657 \\
0.9 0.912485414235706 \\
0.95 0.9393232205367562 \\
1.0 0.9241540256709452 \\
1.05 0.9025670945157526 \\
1.1 0.837222870478413 \\
1.15 0.7771295215869312 \\
1.2 0.822053675612602 \\
1.25 0.0 \\
1.3 0.0 \\
1.35 0.0 \\
1.4 0.0 \\
1.45 0.0 \\
1.5 0.0 \\
1.55 0.0 \\
1.6 0.0 \\
1.65 0.0 \\
1.7 0.0 \\
1.75 0.0 \\
1.8 0.0 \\
1.85 0.0 \\
1.9 0.0 \\
1.95 0.0 \\
2.0 0.0 \\
2.05 0.0 \\
2.1 0.0 \\
2.15 0.0 \\
2.2 0.0 \\
2.25 0.0 \\
2.3 0.0 \\
2.35 0.0 \\
2.4 0.0 \\
2.45 0.0 \\
2.5 0.0 \\
2.55 0.0 \\
2.6 0.0 \\
2.65 0.0 \\
2.7 0.0 \\
2.75 0.0 \\
2.8 0.0 \\
2.85 0.0 \\
2.9 0.0 \\
2.95 0.0 \\
3.0 0.0 \\
3.05 0.0 \\
3.1 0.0 \\
3.15 0.0 \\
3.2 0.0 \\
3.25 0.0 \\
3.3 0.0 \\
3.35 0.0 \\
3.4 0.0 \\
3.45 0.0 \\
3.5 0.0 \\
3.55 0.0 \\
3.6 0.0 \\
3.65 0.0 \\
3.7 0.0 \\
3.75 0.0 \\
3.8 0.0 \\
3.85 0.0 \\
3.9 0.0 \\
3.95 0.0 \\
4.0 0.0 \\
4.05 0.0 \\
4.1 0.0 \\
4.15 0.0 \\
4.2 0.0 \\
4.25 0.0 \\
4.3 0.0 \\
4.35 0.0 \\
4.4 0.0 \\
4.45 0.0 \\
4.5 0.0 \\
4.55 0.0 \\
4.6 0.0 \\
4.65 0.0 \\
4.7 0.0 \\
4.75 0.0 \\
4.8 0.0 \\
0.0 0.5075845974329055 \\
0.05 0.5029171528588098 \\
0.1 0.6306884480746792 \\
0.15 0.7357059509918319 \\
0.2 0.8494749124854143 \\
0.25 0.9667444574095683 \\
0.3 0.9474912485414235 \\
0.35 0.9043173862310385 \\
0.4 0.955659276546091 \\
0.45 0.8786464410735122 \\
0.5 0.9422403733955659 \\
0.55 0.9031505250875146 \\
0.6 0.7485414235705952 \\
0.65 0.5915985997666278 \\
0.7 0.6522753792298717 \\
0.75 0.8022170361726954 \\
0.8 0.9358226371061843 \\
0.85 0.8663943990665111 \\
0.9 0.8745624270711785 \\
0.95 0.9288214702450408 \\
1.0 0.9253208868144691 \\
1.05 0.8984830805134189 \\
1.1 0.853558926487748 \\
1.15 0.7747957992998833 \\
1.2 0.7800466744457409 \\
1.25 0.0 \\
1.3 0.0 \\
1.35 0.0 \\
1.4 0.0 \\
1.45 0.0 \\
1.5 0.0 \\
1.55 0.0 \\
1.6 0.0 \\
1.65 0.0 \\
1.7 0.0 \\
1.75 0.0 \\
1.8 0.0 \\
1.85 0.0 \\
1.9 0.0 \\
1.95 0.0 \\
2.0 0.0 \\
2.05 0.0 \\
2.1 0.0 \\
2.15 0.0 \\
2.2 0.0 \\
2.25 0.0 \\
2.3 0.0 \\
2.35 0.0 \\
2.4 0.0 \\
2.45 0.0 \\
2.5 0.0 \\
2.55 0.0 \\
2.6 0.0 \\
2.65 0.0 \\
2.7 0.0 \\
2.75 0.0 \\
2.8 0.0 \\
2.85 0.0 \\
2.9 0.0 \\
2.95 0.0 \\
3.0 0.0 \\
3.05 0.0 \\
3.1 0.0 \\
3.15 0.0 \\
3.2 0.0 \\
3.25 0.0 \\
3.3 0.0 \\
3.35 0.0 \\
3.4 0.0 \\
3.45 0.0 \\
3.5 0.0 \\
3.55 0.0 \\
3.6 0.0 \\
3.65 0.0 \\
3.7 0.0 \\
3.75 0.0 \\
3.8 0.0 \\
3.85 0.0 \\
3.9 0.0 \\
3.95 0.0 \\
4.0 0.0 \\
4.05 0.0 \\
4.1 0.0 \\
4.15 0.0 \\
4.2 0.0 \\
4.25 0.0 \\
4.3 0.0 \\
4.35 0.0 \\
4.4 0.0 \\
4.45 0.0 \\
4.5 0.0 \\
4.55 0.0 \\
4.6 0.0 \\
4.65 0.0 \\
4.7 0.0 \\
4.75 0.0 \\
4.8 0.0 \\
0.0 0.5997666277712952 \\
0.05 0.6645274212368727 \\
0.1 0.5851808634772462 \\
0.15 0.8033838973162193 \\
0.2 0.9119019836639439 \\
0.25 0.9649941656942823 \\
0.3 0.9346557759626605 \\
0.35 0.808634772462077 \\
0.4 0.9381563593932322 \\
0.45 0.956242707117853 \\
0.5 0.9486581096849475 \\
0.55 0.882147024504084 \\
0.6 0.7438739789964994 \\
0.65 0.4579929988331389 \\
0.7 0.7357059509918319 \\
0.75 0.9025670945157526 \\
0.8 0.7444574095682615 \\
0.85 0.897899649941657 \\
0.9 0.9194865810968494 \\
0.95 0.9422403733955659 \\
1.0 0.8768961493582264 \\
1.05 0.882730455075846 \\
1.1 0.8442240373395566 \\
1.15 0.7987164527421237 \\
1.2 0.8634772462077013 \\
1.25 0.0 \\
1.3 0.0 \\
1.35 0.0 \\
1.4 0.0 \\
1.45 0.0 \\
1.5 0.0 \\
1.55 0.0 \\
1.6 0.0 \\
1.65 0.0 \\
1.7 0.0 \\
1.75 0.0 \\
1.8 0.0 \\
1.85 0.0 \\
1.9 0.0 \\
1.95 0.0 \\
2.0 0.0 \\
2.05 0.0 \\
2.1 0.0 \\
2.15 0.0 \\
2.2 0.0 \\
2.25 0.0 \\
2.3 0.0 \\
2.35 0.0 \\
2.4 0.0 \\
2.45 0.0 \\
2.5 0.0 \\
2.55 0.0 \\
2.6 0.0 \\
2.65 0.0 \\
2.7 0.0 \\
2.75 0.0 \\
2.8 0.0 \\
2.85 0.0 \\
2.9 0.0 \\
2.95 0.0 \\
3.0 0.0 \\
3.05 0.0 \\
3.1 0.0 \\
3.15 0.0 \\
3.2 0.0 \\
3.25 0.0 \\
3.3 0.0 \\
3.35 0.0 \\
3.4 0.0 \\
3.45 0.0 \\
3.5 0.0 \\
3.55 0.0 \\
3.6 0.0 \\
3.65 0.0 \\
3.7 0.0 \\
3.75 0.0 \\
3.8 0.0 \\
3.85 0.0 \\
3.9 0.0 \\
3.95 0.0 \\
4.0 0.0 \\
4.05 0.0 \\
4.1 0.0 \\
4.15 0.0 \\
4.2 0.0 \\
4.25 0.0 \\
4.3 0.0 \\
4.35 0.0 \\
4.4 0.0 \\
4.45 0.0 \\
4.5 0.0 \\
4.55 0.0 \\
4.6 0.0 \\
4.65 0.0 \\
4.7 0.0 \\
4.75 0.0 \\
4.8 0.0 \\
};\addplot [red,dashed,line width=1pt]  coordinates { (1.20,0)(1.20,1.1) };

\nextgroupplot[
xmin = 0, xmax=5.0,
ymin = 0.0, ymax=110.0,
ytick = {0, 50, 100},
ylabel={$S$},
xlabel={Time (\SI{}{\second})},
]
\addplot [color=blue]
table[row sep=crcr]{%
0.0 0.0 \\
0.05 0.0 \\
0.1 0.0 \\
0.15 0.0 \\
0.2 0.0 \\
0.25 0.0 \\
0.3 0.0 \\
0.35 0.0 \\
0.4 0.0 \\
0.45 0.0 \\
0.5 0.0 \\
0.55 0.0 \\
0.6 0.0 \\
0.65 0.0 \\
0.7 0.0 \\
0.75 0.0 \\
0.8 0.0 \\
0.85 0.0 \\
0.9 0.0 \\
0.95 0.0 \\
1.0 0.0 \\
1.05 0.0 \\
1.1 0.0 \\
1.15 0.0 \\
1.2 0.0 \\
1.25 12.41786800206475 \\
1.3 24.8357360041295 \\
1.35 37.25360400619425 \\
1.4 49.671472008259 \\
1.45 62.08934001032375 \\
1.5 74.5072080123885 \\
1.55 86.92507601445325 \\
1.6 12.41786800206475 \\
1.65 24.8357360041295 \\
1.7 37.25360400619425 \\
1.75 49.671472008259 \\
1.8 62.08934001032375 \\
1.85 74.5072080123885 \\
1.9 86.92507601445325 \\
1.95 12.41786800206475 \\
2.0 24.8357360041295 \\
2.05 37.25360400619425 \\
2.1 49.671472008259 \\
2.15 62.08934001032375 \\
2.2 74.5072080123885 \\
2.25 86.92507601445325 \\
2.3 12.41786800206475 \\
2.35 24.8357360041295 \\
2.4 37.25360400619425 \\
2.45 49.671472008259 \\
2.5 62.08934001032375 \\
2.55 74.5072080123885 \\
2.6 86.92507601445325 \\
2.65 12.41786800206475 \\
2.7 24.8357360041295 \\
2.75 37.25360400619425 \\
2.8 49.671472008259 \\
2.85 62.08934001032375 \\
2.9 74.5072080123885 \\
2.95 86.92507601445325 \\
3.0 12.41786800206475 \\
3.05 24.8357360041295 \\
3.1 37.25360400619425 \\
3.15 49.671472008259 \\
3.2 62.08934001032375 \\
3.25 74.5072080123885 \\
3.3 86.92507601445325 \\
3.35 12.41786800206475 \\
3.4 24.8357360041295 \\
3.45 37.25360400619425 \\
3.5 49.671472008259 \\
3.55 62.08934001032375 \\
3.6 74.5072080123885 \\
3.65 86.92507601445325 \\
3.7 12.41786800206475 \\
3.75 24.8357360041295 \\
3.8 37.25360400619425 \\
3.85 49.671472008259 \\
3.9 62.08934001032375 \\
3.95 74.5072080123885 \\
4.0 86.92507601445325 \\
4.05 12.41786800206475 \\
4.1 24.8357360041295 \\
4.15 37.25360400619425 \\
4.2 49.671472008259 \\
4.25 62.08934001032375 \\
4.3 74.5072080123885 \\
4.35 86.92507601445325 \\
4.4 12.41786800206475 \\
4.45 24.8357360041295 \\
4.5 37.25360400619425 \\
4.55 49.671472008259 \\
4.6 62.08934001032375 \\
4.65 74.5072080123885 \\
4.7 86.92507601445325 \\
4.75 12.41786800206475 \\
4.8 24.8357360041295 \\
};\addplot [red,dashed,line width=1pt]  coordinates { (1.20,0)(1.20,220) };

\end{groupplot}
\end{tikzpicture}%

%% file: conclusion.tex
\section{Conclusions}
\label{sec:conclutions}
In this work, we presented a detection method for adversarial examples in learning-enabled CPS. 
The method is based on inductive conformal prediction and uses a VAE-based regression model to predict the target variable and compute the nonconformity of new inputs relative to the training set. 
%The model takes the outputs into the consideration for the adversarial inputs detection.  
The evaluation is based on an AEBS implemented in an open source simulator for self-driving cars. FGSM is used to generate adversarial examples. The results demonstrate that the method can efficiently detect adversarial examples with a short detection delay.  
%\todo{FC: future work}
Future work includes detection of physically realizable attacks, \added{comparing this new approach with other adversarial detection methods to study into the benifit of taking the output into consideration}, and also
investigating the generation of adversarial examples that are not detectable by the approach.